\documentclass[10pt,journal]{IEEEtran}

\ifCLASSINFOpdf
\else
\fi

\usepackage{caption,subcaption}
\usepackage{booktabs}
\usepackage{amsmath}
\usepackage{rotating}
\usepackage[pagebackref=true,breaklinks=true,
letterpaper=true,colorlinks,citecolor=blue,
linkcolor=blue,bookmarks=false]{hyperref}

\usepackage{color}
\usepackage{times}
\usepackage{epsfig}
\usepackage{graphicx}
\usepackage{amssymb}
\usepackage{multirow}
\usepackage{algorithm}
\usepackage{algorithmic}
\usepackage{array}
\usepackage{cite}
\newcommand{\RN}[1]{%
  \textup{\uppercase\expandafter{\romannumeral#1}}%
}

\begin{document}

\title{Muti-view Mouse Social Behaviour Recognition with Deep Graphic Model}

\author{Zheheng Jiang*, Feixiang Zhou*, Aite Zhao, Xin Li, Ling Li, Dacheng Tao,~\IEEEmembership{Fellow,~IEEE}, Xuelong Li,~\IEEEmembership{Fellow,~IEEE} and Huiyu Zhou% <-this % stops a space
\thanks{Z. Jiang is with School of Computing and Communications, Lancaster University, United Kingdom. E-mail: z.jiang11@lancaster.ac.uk.}
\thanks{F. Zhou and H. Zhou are with School of Informatics, University of Leicester, United Kingdom. E-mail: \{fz64;hz143\}@leicester.ac.uk. Z. Jiang and F. Zhou contributed equally to the study. H. Zhou is the corresponding author. H. Zhou is supported in part by Royal Society-Newton Advanced Fellowship under Grant NA160342.}
\thanks{X. Li is with School of Engineering and Department of Cardiovascular Sciences, University of Leicester, United Kingdom}
\thanks{A. Zhao is with Department of information science and engineering, Ocean University of China, Qingdao, 266100, China.}
\thanks{L. Li is with the School of Computing, University of Kent, United Kingdom.}
\thanks{D. Tao is with the JD Explore Academy in JD.com, China. E-mail: dacheng.tao@gmail.com.}
\thanks{X. Li is with School of Artificial Intelligence, Optics and Electronics (iOPEN), Northwestern Polytechnical University, Xi'an 710072, P.R. China. E-mail: xuelong\_li@nwpu.edu.cn.}
% <-this % stops a space

\thanks{Manuscript submitted in September 2020; revised xxxx.}}

\maketitle

\begin{abstract}
Home-cage social behaviour analysis of mice is an invaluable tool to assess therapeutic efficacy of neurodegenerative diseases. Despite tremendous efforts made within the research community, single-camera video recordings are mainly used for such analysis. Because of the potential to create rich descriptions for mouse social behaviors, the use of multi-view video recordings for rodent observations is increasingly receiving much attention. However, identifying social behaviours from various views is still challenging due to the lack of correspondence across data sources. To address this problem, we here propose a novel multi-view latent-attention and dynamic discriminative model that jointly learns view-specific and view-shared sub-structures, where the former captures unique dynamics of each view whilst the latter encodes the interaction between the views. Furthermore, a novel multi-view latent-attention variational autoencoder model is introduced in learning the acquired features, enabling us to learn discriminative features in each view. Experimental results on the standard CRMI13 and our multi-view Parkinson's Disease Mouse Behaviour (PDMB) datasets demonstrate that our proposed model outperforms the other state of the arts technologies, has lower computational cost than the other graphical models and effectively deals with the imbalanced data problem.
\end{abstract}

\IEEEpeerreviewmaketitle

\section{Introduction}
Mouse models have been extensively developed to study across cognitive and neurological fields for Down syndrome\cite{olson2004down}, autism\cite{penagarikano2011absence}, Alzheimer's disease\cite{lewejohann2009behavioral} and Parkinson's disease\cite{blume2009stepping}. Comprehensive behavioural phenotypes of transgenic mice can be used to reveal the underlying functional role of genes, and provide new insights into the pathophysiology and treatment of the diseases carried by the mice\cite{iancu2005behavioral,montkowski1995long,
kilpelainen2019behavioural,liu2016autism}. Historically, such behaviour is primarily labelled by a human expert, which is a time-consuming, labor-intensive and error-prone task. To reduce the inherent high labour cost and inter-investigator variability associated with the manual annotation of data, reliable and high-throughput methods for automated quantitative analysis of mouse behaviours have become extremely important.

Previous automated systems have mainly relied on the use of various sensors to monitor animal behaviours. These established technologies include the use of infrared sensors\cite{casadesus2001automated}, radio-frequency identification (RFID) transponders\cite{lewejohann2009behavioral} and photobeams\cite{goulding2008robust}. Such approaches have been successfully applied to the analysis of simple pre-programmed behaviours such as running and resting. However, the capacity of these sensor-based approaches restricts the complexity of the objects’ behaviours that can be measured. They cannot be used to handle more complex mouse behaviours such as eating, attacking, or sniffing. Vision-based techniques is thus used to recognise subtle mouse behaviours.

\begin{table*}
\begin{center}
\small
\caption{Ethogram of the observed behaviours, derived from CRIM13\cite{burgos2012social}.}
\label{tab:CRIM13_description}
\begin{tabular}{|p{2cm}|p{15cm}|}
\hline
Behaviour & Description\\
\hline\hline
approach & Moving toward another mouse in a straight line without obvious exploration\\
attack & Biting/pulling fur of another mouse\\
copulation  & Copulation of male and female mice\\
chase & A following mouse attempts to maintain a close distance to another mouse while the latter is moving.\\
circle & Circling around own axis or chasing tail\\
drink & Licking at the spout of the water bottle\\
eat & Gnawing/eating food pellets held by the fore-paws\\
clean & Washing the muzzle with fore-paws (including licking fore-paws) or grooming the fur or hind-paws by means of licking or chewing\\
human & Human intervenes with mice\\
sniff & Sniff any body part of another mouse\\
up & Exploring while standing in an upright posture\\
walk away & Moving away from another mouse in a straight line without obvious exploration\\
other & Behaviour other than defined in this ethogram, or when it is not visible what behaviour the mouse displays\\
\hline
\end{tabular}
\end{center}
\end{table*}

\begin{figure*}
\begin{center}
\fbox{\includegraphics[width=18cm]{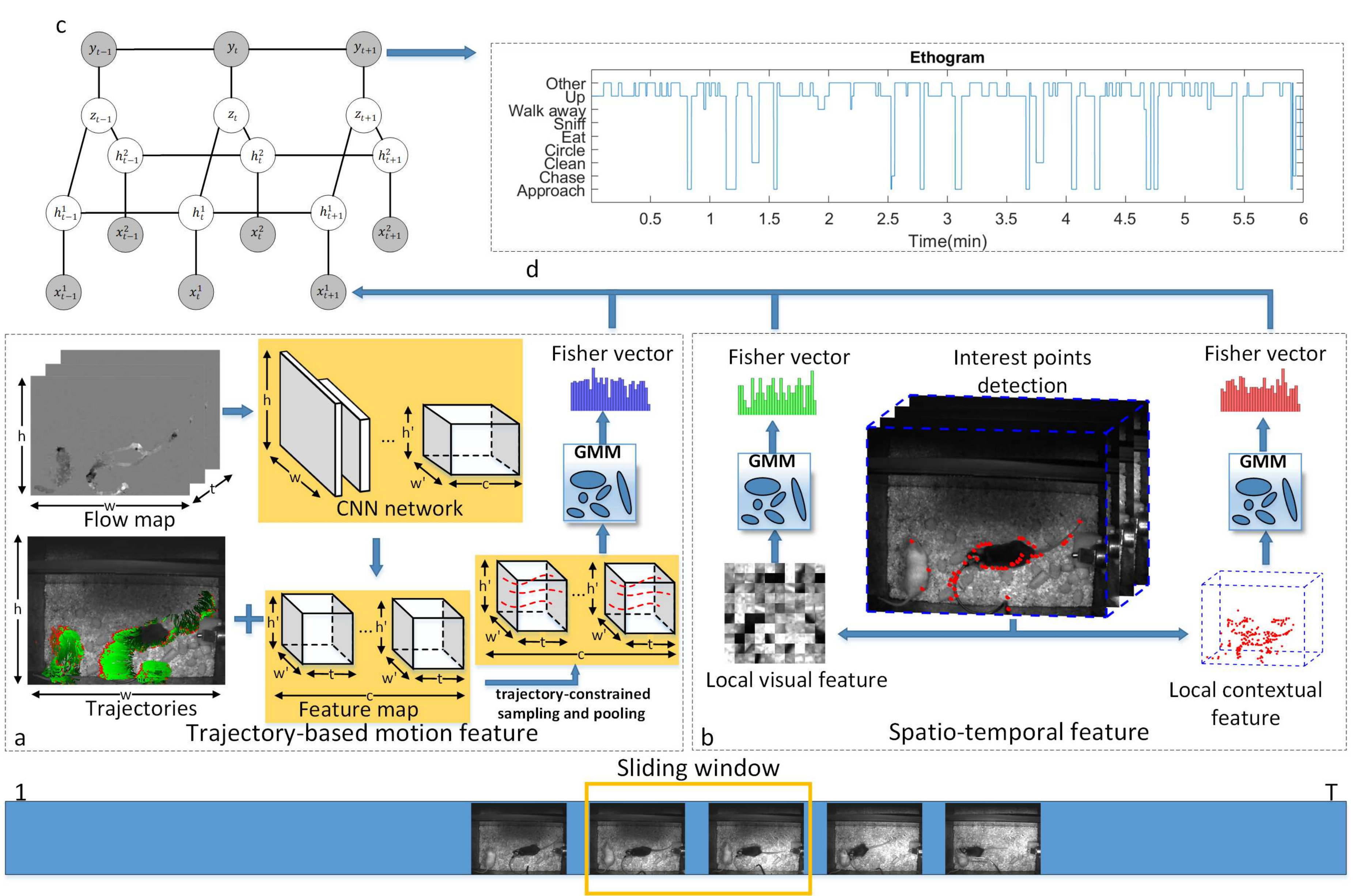}}
\end{center}
\caption{Overview of the proposed system for multi-view mouse behaviour recognition. There are two types of features (a) and (b) that are computed in sliding windows centered at each frame. (a) For the computation of trajectory-based motion features, a set of points are densely sampled at each frame. After having eliminated points in homogeneous areas by setting a threshold to drop the smaller eigenvalue of their autocorrelation matrices, the remaining points are then tracked by deploying median filtering in a dense flow field. To efficiently depict the tracked points, following \cite{wang2015action}, we conduct trajectory-constrained sampling and pooling over convolutional feature maps, based on optical flows, to retain trajectory-pooled deep convolutional descriptors. (b) To extract spatio-temporal features, spatio-temporal interest points are first generated by applying a Laplacian of Gaussian (LoG) filter along the spatial dimension and a quadrature pair of 1-D Gabor filters along the temporal dimension. Two types of local features are then computed: local visual and contextual features. More details about (a) and (b) can be found in Section 2.2. To efficiently fuse features extracted in (a) and (b), which are depicted in different feature spaces, we apply Fisher Vectors (FV) \cite{sanchez2013image} with Gaussian Mixture Models to encoding the features. (c) Our proposed Multi-view Latent-Attention and Dynamic Discriminative Model, where each node $x^{v}_{t}$ models the input feature computed from different features of the $v_{th}$ camera view at timestamp $t$, $h_{t}^{v}$ models the view-specific sub-structure and $z_{t}$ models the deep view-shared sub-structure (detailed in Section 2.3). At the same time, the use of the FV technique can ensure all $x^{v}_{t}$ nodes to be represented in the same feature space. (d) An ethogram illustrates the sequence of the labels predicted by our proposed model.}
\label{fig:framework}
\end{figure*}

Benefiting from the advances made in computer vision and machine learning over the last decade, several vision-based approaches for automated tracking\cite{hong2015automated,de2012computerized,ohayon2013automated} and recognition of mouse behaviours \cite{van2013automated,jhuang2010automated,robie2017machine,jiang2018context} have been constructed. However, most of them rely on the analysis of single-view video recordings, which can be ambiguous when essential information of behaviours is occluded. In this paper, we are particularly interested in recognising mouse behaviours (see Table \ref{tab:CRIM13_description} for the description of mouse behaviours) by using multi-view video recordings, which is a challenging task due to large data variations over different views. 

Probabilistic graphical models are a useful tool to address the dynamic behaviour recognition problem due to their ability in fully exploiting spatial and temporal structures of data\cite{baltruvsaitis2018multimodal}. Normally, graphical models can be classified into two main categories: generative and discriminative models\cite{li2018survey}. Some of the popular approaches use generative models such as Hidden Markov Model (HMM) and Dynamic Bayesian Networks. In particular, Brand et al.\cite{brand1997coupled} introduced a coupled HMM to model interacting processes, and Murphy et al.\cite{murphy2002dynamic} introduced Dynamic Bayesian Networks to model complex dependencies in the hidden (or observed) state variables. Comparatively, discriminative models such as conditional random fields (CRFs) are more commonly used due to their better predictive power than the generative ones\cite{li2018survey}. CRFs have been extended to model the latent states, e.g. using Hidden Conditional Random Field (HCRF) \cite{quattoni2007hidden}. Latent Discriminative HCRFs (LDCRF) \cite{morency2007latent} is a variation of HCRF tailored to deal with the dynamic behaviour recognition problem. Song et al.\cite{song2012multi} further extended LDCRF to the multi-view (MV) domain and proposed a MV-LDCRF model by defining view-specific and view-shared edges. Our work is also based on graphical model due to its advantages of representing and reasoning over structured data. However, different from the above graphical models, we integrate a deep neural network and a graphical model to resolve view-specific and view-shared features learning problems by proposing a novel multi-view latent-attention variational autoencoder model. Moreover, our graphical model also model the correlation between the neighbouring labels, which has shown superior performance to recognise mouse behaviours in a long video recording.

In this paper, we describe a novel multi-view mouse behaviour recognition system based on trajectory-based motion and spatio-temporal features as shown in Fig. \ref{fig:framework}. Specifically, we here propose a novel deep probabilistic graphical model with the aims to model: (1) the temporal relationship of image frames in each view, (2) the relationship between camera views, and (3) the correlations between the neighbouring labels.

\section{Related Work}
\subsection{Mouse Behaviour Recognition}
In the literature, several open-source and commercial computer vision systems have also
been developed to recognise mouse behaviours. For instance, de Chaumont et al.\cite{de2012computerized} and Giancardo et al.\cite{giancardo2013automatic} firstly estimated the positions of the mouse body parts (e.g. head and trunk) by deploying a geometrical primitive model and a temporal watershed segmentation algorithm respectively, and then recognised mouse behaviours based on these positions. Since they only use top-view video recordings, it is difficult to recognise some behaviours that involve vertical movements e.g. `rearing'. In contrast, the side-view video recordings may supply a better perspective for bouts of behaviour. For example, Jhuang et al.\cite{jhuang2010automated} extracted biologically inspired features from the side view, followed by classification using a Hidden Markov Model Support Vector Machine (SVMHMM) method. Jiang et al.\cite{jiang2018context} developed and implemented a novel Hidden Markov Model algorithm for behaviour recognition using visual and contextual features. These systems were successful for measuring single mouse behaviour. If multiple mice are in the scene, these systems lack the ability to recognise the interactions between mice due to occlusion or clutters. In such case, the ambiguity caused by occlusion can be mitigated by adopting multiple-view observations. Burgos-Artizzu et al.\cite{burgos2012social} designed a system for recognising social behaviours of a mouse from both top and side views. They firstly extracted spatio-temporal and trajectory features and then applied AdaBoost to classifying those extracted features. However, their approach can only learn view-specific feature representations. The relationship between different camera views and the temporal transition of mouse behaviours are not addressed in their approach. Hong et al.\cite{hong2015automated} utilised a top-view camera and a top-view depth sensor to track and extract the body-pose features of mice by fitting an ellipse to each of them. These body pose features are then integrated with pixel changes from the side-view to train a classifier. Similar to the method of Burgos-Artizzu et al. \cite{burgos2012social}, their method also ignored the relationship between different camera views and the temporal transition of mouse behaviours.  Another popular method employing multi-view cameras is to reconstruct the 3D pose of a mouse\cite{matsumoto20133d,sheets2013quantitative,
salem2019three}, but it requires additional equipment, calibration of cameras, computational resources, and 3D tracking software.
\subsection{Human Behaviour Recognition}
Human individual behaviour and group activity recognition have also attracted large research interests in the community of computer vision. It is commonly formulated as a classification problem over a short video segment of a few seconds. Recently, deep learning based approaches are widely used to extract feature from video segments. For instance, Simonyan et al. \cite{simonyan2014two} proposed a two-stream CNN architecture to learn representations respectively from input RGB images and optical flows. Wang et al. \cite{wang2015action} designed a trajectory-pooled deep convolutional descriptor (TDD) for combining the benefits of both trajectory-based and deep-learned features. In order to capture relevant relation between actors for group activity recognition, several works firstly detect and track actors in the video and then modelled the relationship between the actors based on graphical models \cite{choi2012unified, shu2015joint,ibrahim2016hierarchical,shu2017cern,
wu2019learning}. However, these models are  computationally costly and their performance is sensitive to the human detector and tracker. Moreover, features directly extracted from detector or tracker are sometimes ambiguous in contact and occlusion situations. This issue can be alleviated by installing multiple cameras at different view points. Recently, several approaches have been proposed to address the problem of multi-view action recognition. Liu et al. \cite{liu2011cross} presented a bipartite-graph-based method to bridge the semantic gap across view-dependent vocabularies. Zheng et al. \cite{zheng2016cross} proposed to learn a set of view-specific dictionaries for individual views and a common dictionary can be shared by different views. Junejo et al. \cite{junejo2008cross} summarised actions at various views using a so-called self-similarity matrix (SSM) descriptor. In order to enhance the representation power of SSM, Yan et al. \cite{yan2014multitask} proposed a multi-task learning approach to share discriminative SSM features between different views. However, these methods can only deal with segmented sequences, each of which contains only one subject's behaviours.

\subsection{Comparisons Between Human and Mouse Behaviour Recognition}
Although the tasks of human and mouse behaviour recognition interestingly share a few basic concepts, they have different requirements and challenges which we want to elaborate
in this section. First, most existing human behaviour recognition methods focus on classification of short video segments, which generally last for several seconds, such as UCF-101 \cite{soomro2012ucf101} and Volleyball Dataset \cite{ramanathan2016detecting}. Very few human behaviour recognition methods attempt to model behavioural label correlation that is very important to support the recognition of mouse behaviours in a long video recording. Second, different from most human subject datasets, the behaviours in the mouse dataset are highly imbalanced. For example, the majority (56\%) of the CRIM13 dataset is labelled as `other' while `drink' only has `0.4\%' of the whole dataset. Such an imbalance poses certain challenges to mouse behaviour recognition methods in both training and prediction.

\section{Proposed Methods}
In this section, we give full details to our proposed feature extraction approach that extracts discriminative features from videos, and our proposed MV-LADDM model that fuses and dynamically classifies these extracted features. The overview of the proposed system is illustrated in Fig. \ref{fig:framework}.
\subsection{Feature Extraction}
From the video data, two types of features were extracted: spatio-temporal features and trajectory-based motion features. Each of these features was rigorously chosen to capture different aspects of the mouse posture and movement. The spatio-temporal features used in this study include local visual and contextual features. Both of them are based on the extracted spatio-temporal interest points, obtained by employing a Laplacian of Gaussian (LoG) filter along the spatial dimension and a quadrature pair of 1-D Gabor  filters along the temporal dimension. For the computation of local visual features, we extract
the brightness gradients of three channels $\left ( G_{x},G_{y},G_{z} \right )$ from the cuboid of each interest point. The contextual features can be computed in the form: $F_{q}=\frac{\left [ X_{q}-X_{c};Y_{q}-Y_{c};X_{q};Y_{q} \right ]}{\left \| \left [ X_{q}-X_{c};Y_{q}-Y_{c};X_{q};Y_{q} \right ] \right \|_{2}},q=1,2,...,Q$ where $\left [ X_{c};Y_{c};T_{c} \right ]$ and $\left [ X_{q};Y_{q};T_{q} \right ]$ represent the coordinates of the centre and the $q$th interest point respectively\cite{jiang2018context}. These features can characterise both the spatial location and temporal changes of mice.

Trajectory-based motion features\cite{wang2015action} are the combination of dense trajectories and deeply learned features since deep learning has produced remarkable progress in human action recognition\cite{wang2015action,simonyan2014two,
antonik2019human,pang2020complex}. The first step of computing dense trajectories is to densely sample a set of points on a grid with the step size of 5 pixels on 8 spatial scales, which has been justified to produce satisfactory results in \cite{wang2011action}. Points in homogeneous areas are eliminated if the eigenvalues of their autocorrelation matrices are below a pre-defined threshold. Afterwards, these sampled points are tracked using a median filter in a dense flow field. To generate deeply learned features, we adopt the temporal stream nets proposed in \cite{simonyan2014two}. The temporal stream nets are trained on the stacking optical flow field of the action dataset, describing the dynamic motion information. Similar to \cite{wang2015action}, we also choose the trajectory-constrained sampling and pooling descriptors from conv3 and con4 layers of the temporal stream nets. Finally, we de-correlate TDD by PCA and reduce its dimensionality.

We apply Fisher Vectors (FVs) \cite{sanchez2013image} to encoding all the features into high dimensional representations that have been proved to be effective for action recognition in previous works \cite{wang2013action,wang2015action,jiang2018context}.
We firstly train a Gaussian Mixture Model (GMM) with parameters $\lambda =\left \{ \omega_{k},\mu_{k},\sigma_{k},k=1,...,K \right \}$ for each type of features. Here, $\omega_{k},\mu_{k},\sigma_{k}$ and $K\left(K=50\right)$ respectively denote the mixture weight, mean vector, standard deviation vector (diagonal covariance) and the number of Gaussians. Then, FV can be computed in the following form:
\begin{equation}
\small
\mathcal{G}_{\mu ,k}^{X}=\frac{1}{N\sqrt{\omega _{k}}}\sum_{n=1}^{N}\gamma _{n}\left ( k \right )\left ( \frac{x_{n}-\mu _{k}}{\sigma _{k}} \right )
\label{equation:G_a}
\end{equation}
\begin{equation}
\small
\mathcal{G}_{\sigma  ,k}^{X}=\frac{1}{N\sqrt{\omega _{k}}}\sum_{n=1}^{N}\gamma _{n}\left ( k \right )\left [ \frac{\left (x_{n}-\mu _{k}\right )^{2}}{\sigma _{k}^{2}}-1 \right ]
\label{equation:G_b}
\end{equation}
where $N$ is the number of the interest points or trajectories within a sliding window. Parameter $\gamma _{n}\left ( k \right )$ is the weight of $x _{n}$ to the $k$th Gaussian: $\gamma_{n} \left ( k \right )=\frac{\omega_{k}u_{k}\left ( x_{n};\mu_{k},\Sigma_{k} \right )}{\sum_{k=1}^{K}\omega_{k}u_{k}\left ( x_{n};\mu_{k},\Sigma_{k} \right )}$. We concatenate $\mathcal{G}_{\mu ,k}^{X}$ and $\mathcal{G}_{\sigma  ,k}^{X}$ after having used power normalisation, followed by $\mathit{L_2}$ normalisation to each of them. Finally, we create a view-specific feature for each sliding window concatenating the FVs computed from all the features as shown in Fig. \ref{fig:framework}.

\begin{figure*}
\begin{center}
\begin{tabular}{ccc}
\begin{subfigure}[b]{0.3\linewidth}
\centering\includegraphics[width=4.2cm]{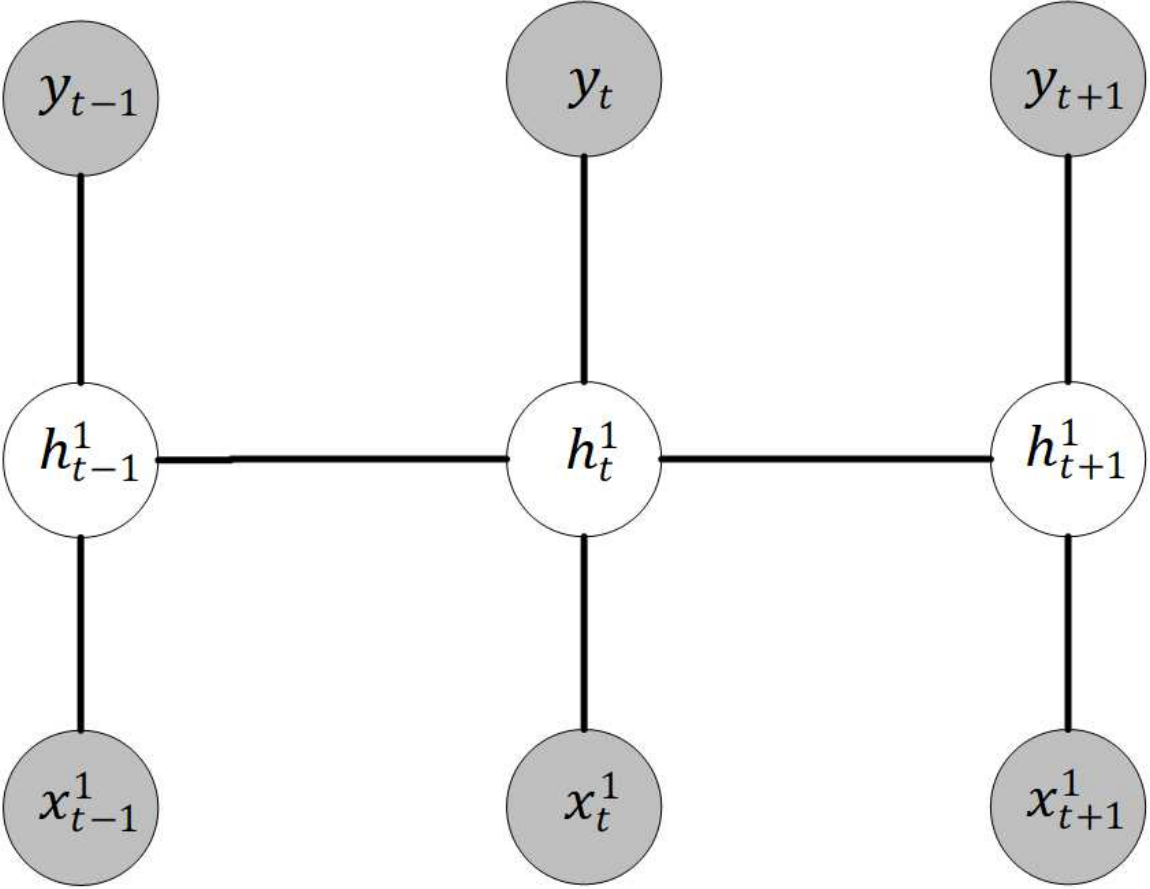}
\subcaption{\label{fig:LDCRF}LDCRF}
\end{subfigure}&
\begin{subfigure}[b]{0.3\linewidth}
\centering\includegraphics[width=5.6cm]{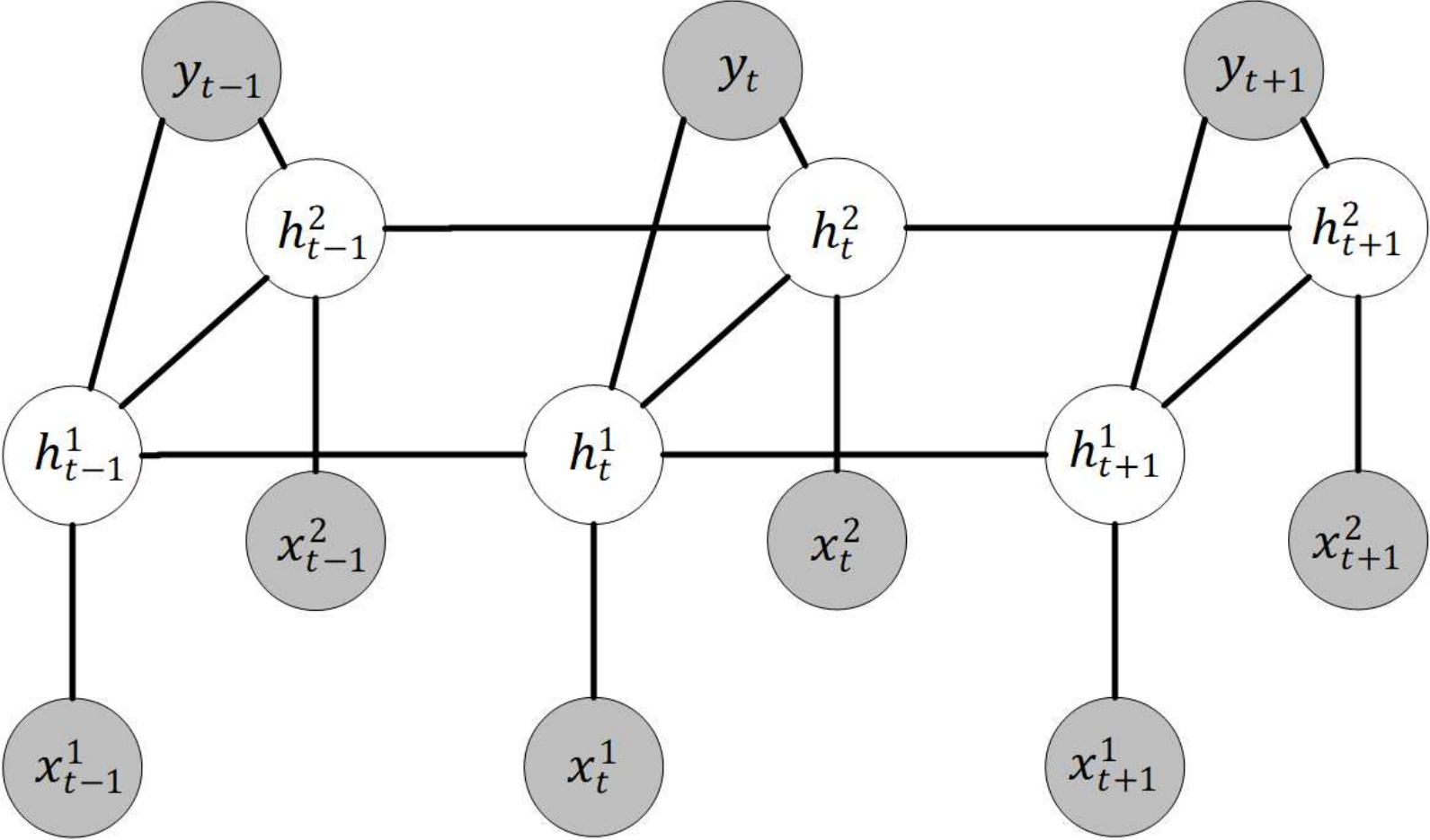}
\subcaption{\label{fig:MV_LDCRF}MV-LDCRF}
\end{subfigure}&
\begin{subfigure}[b]{0.3\linewidth}
\centering\includegraphics[width=5.8cm]{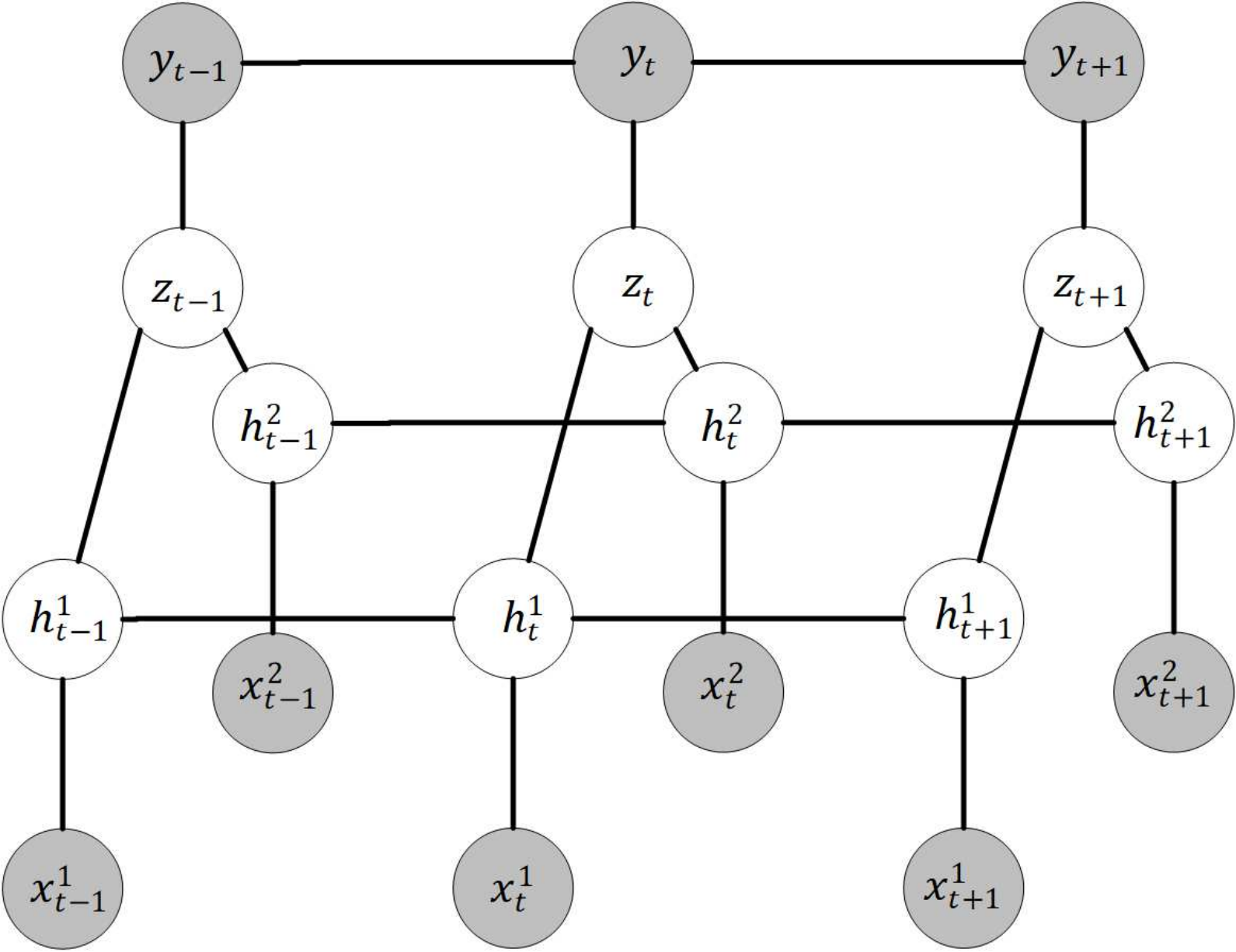}
\subcaption{\label{fig:MV-LADDM}MV-LADDM}
\end{subfigure}
\end{tabular}
\end{center}
\caption{Comparison of our MV-LADDM with two established models: LDCRF\cite{morency2007latent} and MV-LDCRF\cite{song2012multi}. Grey circles are the observed variables and white circles are the latent variables. In these published graphical models, $x^{v}_{t}$ represent the features extracted from the $v_{th}$ view at the timestamp $t$, $h_{t}^{t}$ and $z_{t}$ are the hidden nodes assigned to $x^{v}_{t}$, and $y_{t}$ is the behaviour label at the timestamp $t$. LDCRF is a single-view latent variable discriminative model. MV-LDCRF extends the work of LDCRF to a multi-view domain, but ignores the correlations between the neighbouring labels and is not sufficient to learn a high level of knowledge representations. For the comparison, we introduce a set of higher level latent variables ($z_{t}$) for the deep view-shared representation. Considering the strong dependency across the output labels, we add edges between the neighbouring labels for encoding the temporal transition of social behaviours. Note that we here only illustrate a two-view model for simplicity but our model can be easily generalised to $\geq$ 2 views.}
\label{fig:graphical_representations}
\end{figure*}

\subsection{Multi-view Latent-Attention Dynamic Discriminative Model}
In our model, we denote the input as a set of multi-view sequences $X=\{x^{1},...,x^{V}\}$, where each $x^{v}$ consists of an observation sequence $\{x_{1}^{v},...,x_{T}^{v}\}$ of length $T$ from the $v$-th view. Each $x_{t}$ is associated with a label $y_{t}\in \mathcal{Y}$ at the timestamp $t$. Similar to MV-LDCRF\cite{song2012multi} which extends LDCRF\cite{morency2007latent} (as shown in Fig.\ref{fig:LDCRF}) to model the sub-structure of the multi-view sequences, we also use latent variables. However, different from their methods, where the hidden variables are contemporaneously connected between views as shown in Fig. \ref{fig:MV_LDCRF}, we instead introduce a set of higher level latent variables for deep view-shared representations. In addition, since there are strong dependency across the output labels, for example, social behaviours often switches back and forth between ‘approach’ and ‘walk away’ in our test videos, we add edges between the neighbouring labels for encoding the temporal transition of social behaviours as shown in Fig. \ref{fig:MV-LADDM}. Let $H=\{h^{1},...,h^{V}\}$, where each $h^{v}=\{h_{1}^{v},...,h_{T}^{v}\}$ is a hidden state sequence of length $T$, modelling the view-specific sub-structure, and $Z=\{z_{1},...,z_{T}\}$ models the deep view-shared sub-structure. We are interested in modelling the conditional probability $p(Y|X, \Theta)$ parameterised by $\Theta$, where $Y=\{y_{1},...,y_{t}\}$ is a sequence of labels. The conditional distribution with latent variables $Z$ and $H$ can be modeled as follows:

\begin{equation}
\small
\begin{split}
p\left(Y|X,\Theta\right)&=\sum_{Z}P\left(Y|Z,X,\Theta\right)P\left(Z|X,\Theta\right)\\&=\sum_{Z}\left(\sum_{H}P\left(Y|Z,H,X,\Theta\right)P\left(H|Z,X,\Theta\right)\right.\\&\left.\quad\sum_{H}P\left(Z|H,X,\Theta\right)P\left(H|X,\Theta\right)\right)
\label{equation:cond_distribution1}
\end{split}
\end{equation}

To describe the relationship between random variables, we represent our model as a Markov random field or undirected graph $\mathcal{G}=(\mathcal{V},\mathcal{E})$, where $\mathcal{V}= Y\cup Z\cup H \cup X$ and $\mathcal{E}=\mathcal{E}_{Y} \cup \mathcal{E}_{YZ} \cup \mathcal{E}_{ZH} \cup \mathcal{E}_{H} \cup \mathcal{E}_{HX}$. $\mathcal{E}_{Y}, \mathcal{E}_{YZ}, \mathcal{E}_{ZH}, \mathcal{E}_{H}$ and $\mathcal{E}_{HX}$ denote a set of edges connecting labels, connecting view-sharing latent variables $Z$ with view-specific latent variables $H$, whilst connecting view-specific latent variables $H$ and connecting view-specific latent variables $H$ with observation sequences $X$. Based on the global Markov property, variables $Y$ and $H$ are conditionally independent given variables $Z$, shown in Fig. \ref{fig:MV-LADDM}. We also observe that variables $X$ and $\{Y,Z\}$ are conditionally independent given variables $H$. Hence, we can express our model as follows:

\begin{equation}
\small
\begin{split}
p\left(Y|X,\Theta\right)&=\sum_{Z}\left(\sum_{H}P\left(Y|Z,H,\Theta\right)P\left(H|Z,\Theta\right)\right.\\&\left.\quad\sum_{H}P\left(Z|H,X,\Theta\right)P\left(H|X,\Theta\right)\right)\\&=\sum_{Z}P\left(Y|Z,\Theta\right)\sum_{H}P\left(Z|H,\Theta\right)P\left(H|X,\Theta\right)\\&=\sum_{Z}\sum_{H}P\left(Y|Z,\Theta\right)P\left(Z|H,\Theta\right)P\left(H|X,\Theta\right)
\label{equation:cond_distribution2}
\end{split}
\end{equation}

Eq. (\ref{equation:cond_distribution2}) can be characterised by the Gibbs distribution \cite{li1994markov}:
\begin{equation}
\small
\begin{split}
p\left(Y|X,\Theta\right)=&\sum_{Z}\sum_{H}\frac{e^{-En\left(Y,Z,\Theta\right) }}{\mathcal{Z}\left(Y\right)}\frac{e^{-En\left(Z,H,\Theta\right) }}{\mathcal{Z}\left(Z\right)}\frac{e^{-En\left(H,X,\Theta\right) }}{\mathcal{Z}\left(H\right)}
\label{equation:cond_distribution3}
\end{split}
\end{equation}

$En\left(Y,Z,\Theta\right)$, $En\left(Z,H,\Theta\right)$ and $En\left(H,X,\Theta\right)$ are energy functions to be defined later. $\mathcal{Z}\left(Y\right)=\sum_{Y}e^{-En\left(Y,Z,\Theta\right)}$, $\mathcal{Z}\left(Z\right)=\sum_{Z}e^{-En\left(Z,H,\Theta\right)}$ and $\mathcal{Z}\left(H\right)=\sum_{H}e^{-En\left(H,X,\Theta\right)}$ are partition functions for normalisation.

\subsubsection{Energy functions}
Similar to \cite{song2012multi,quattoni2007hidden}, our energy functions are dependant on how edges $\mathcal{E}$ are defined. The energy function $En\left(Y,Z,\Theta\right)$ in our model is factorised as follows:

\begin{equation}
\small
\begin{split}
En\left(Y,Z,\Theta\right)=\sum_{t} \mathcal{E}_{Y}\left(y_{t-1},y_{t} \right )+\sum_{i}\mathcal{E}_{YZ}\left(y_{t},z_{t} \right )
\label{equation:En_YZT}
\end{split}
\end{equation}
where $\mathcal{E}_{Y}\left(\cdot \right )$ and $\mathcal{E}_{YZ}\left(\cdot\right )$ are two feature functions defined on edges $\mathcal{E}_{Y}$ and $\mathcal{E}_{YZ}$, which encode the relationship between the neighbouring labels and between variables Y and Z, respectively. We represent $\mathcal{E}_{Y}\left(\cdot \right )$ as a $N \times N$ transition score matrix $B\in\Theta$, where $N$ is the number of behaviours shown in Table \ref{tab:CRIM13_description}. Each element $b_{nn^{'}}$ of $B$ denotes the transition score from labels $b_{n}$ to $b_{n^{'}}$ in the next timestamp, i.e. $\mathcal{E}_{Y}\left(y_{t-1}=b_{n},y_{t}=b_{n^{'}} \right )=-b_{nn^{'}}$ and $b_{n}$, $b_{n^{'}}\in \mathcal{Y}$. $\mathcal{E}_{YZ}\left(y_{t},z_{t}\right )$ is represented as $-W_{z_{t},y_{t}}z_{t}$, where $W_{z_{t},y_{t}}\in\Theta$ is the weight vector and the inner product of $W_{z_{t},y_{t}}z_{t}$ can be interpreted as a measure of the plausibility of label $y_{t}$ given $z_{t}$.

\begin{figure}
\begin{center}
\begin{tabular}{c}
\includegraphics[width=7.5cm]{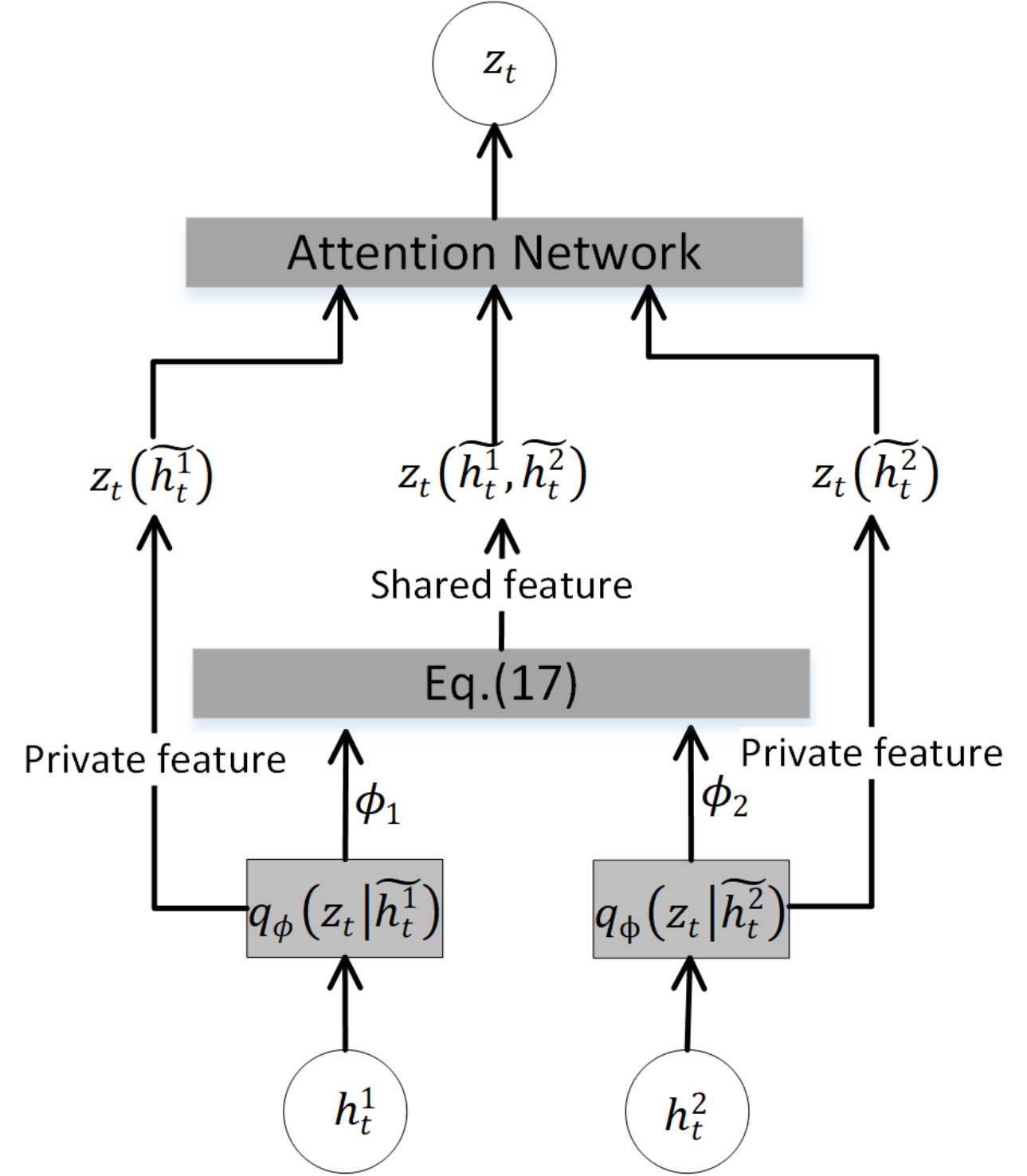}\\
\end{tabular}
\end{center}
\caption{MLVAE architecture with two views. $q_{\phi}\left(z_{t}|\widetilde{h}_{t}^{v}\right)$ represents the inference network with parameters $\phi \in \Theta$ of the $v_{th}$ view. Eq. (\ref{equation:Q_z_1}) combines all the variational parameters in a principled and efficient manner. The attention network can learn attention weights for both view-shared and view-specific latent variables. This architecture is flexible and can be extended for more views.}
\label{fig:MLVAE}
\end{figure}

The energy function $En\left(Z,H,\Theta\right)$ for our model is:
\begin{equation}
\small
\begin{split}
En\left(Z,H,\Theta\right)=\sum_{t}\mathcal{E}_{ZH}\left(z_{t},h_{t}^{1},h_{t}^{2},...,h_{t}^{V}\right)
\label{equation:En_ZHT}
\end{split}
\end{equation}
where $\mathcal{E}_{ZH}$ encodes the relationship between variables $Z$ and $H$. In $En\left(Z,H,\Theta\right)$, we assume the hidden states $h_{t}^{1},...,h_{t}^{V}$ from $V$ views are conditionally independent, given the latent variable $Z$. The latent variable $Z$ is used to represent multi-view data. A common probabilistic graphical model to represent multi-view data is deep Boltzmann machines (DBM) \cite{salakhutdinov2009deep} that stacks the restricted Boltzmann machines (RBM)\cite{hinton2006fast} as building blocks. However, as described in \cite{hinton2012practical}, the latent variable $Z$ is preferred to be binary when we use RBMs. If both variables $Z$ and $H$ are Gaussian, the instability in training RBM becomes worse \cite{hinton2006fast}. Moreover, it is computationally expensive to train RBM using high-dimensional data because of the Monte Carlo practice. Recently, variational autoencoders (VAEs) \cite{kingma2013auto} have been proposed to overcome the above challenging problems. However, how to extend VAE for handling multi-view data is still an open challenge. Here, we introduce a multi-view latent-attention variational autoencoder (MLVAE) (see Fig. \ref{fig:MLVAE}) that uses a multi-Gaussian inference model in combination with latent attention networks to solve the multi-view inference problem.

Since Eq. (\ref{equation:cond_distribution3}) needs to marginalise latent variables $Z$ and $H$ and derive $p\left(Y|X,\Theta\right)$, its computational complexity is exponentially proportional to the cardinality of $Z$ and $H$. To infer $p\left(Y|X,\Theta\right)$ in an efficient way, following the approximation used in greedy layer-wise learning for deep belief nets reported in \cite{hinton2006fast}, we formulate:
\begin{equation}
\small
\begin{split}
p\left(Y|X,\Theta\right)&=\sum_{Z}P\left(Y|Z,X,\Theta\right)P\left(Z|X,\Theta\right)\\&\approx P\left(Y|\widetilde{Z},\Theta\right)
\label{equation:approximation_Y}
\end{split}
\end{equation}

\begin{equation}
\small
\begin{split}
where\quad &\widetilde{Z}=\left\{\widetilde{z}_{1},...,\widetilde{z}_{T}\right\}=\left\{E\left[z_{1}\right],...,E\left[z_{T}\right]\right\}\\&E\left[z_{t}\right]=\sum_{z_{t}}z_{t}P\left(z_{t}|X,\Theta\right)
\label{equation:average_Z}
\end{split}
\end{equation}

\begin{equation}
\small
\begin{split}
P\left(z_{t}|X,\Theta\right)&=\sum_{h_{t}}P\left(z_{t}|h_{t},X,\Theta\right)P\left(h_{t}|X,\Theta\right)\\&\approx P\left(z_{t}|\widetilde{h}_{t},\Theta\right)
\label{equation:approximation_Z}
\end{split}
\end{equation}

\begin{equation}
\small
\begin{split}
where\quad &\widetilde{h}_{t}=\left\{\widetilde{h}_{t}^{1},...,\widetilde{h}_{t}^{V}\right\}=\left\{E\left[h_{t}^{1}\right],...,E\left[h_{t}^{V}\right]\right\}\\&E\left[h_{t}^{v}\right]=\sum_{h_{t}^{v}}h_{t}^{v}P\left(h_{t}^{v}|x^{v},\Theta\right)
\label{equation:average_h}
\end{split}
\end{equation}
In Eqs. (\ref{equation:approximation_Y}), (\ref{equation:average_Z}), (\ref{equation:approximation_Z}) and (\ref{equation:average_h}), we replace $Z$ and $h_{t}$ by their averaging configuration $\widetilde{Z}=\left\{E\left[z_{1}\right],...,E\left[z_{T}\right]\right\}$ and $\widetilde{h}_{t}=\left\{E\left[h_{t}^{1}\right],...,E\left[h_{t}^{V}\right]\right\}$.

Eq. (\ref{equation:En_YZT}) can be used to derive the probability $P\left(Y|\widetilde{Z},\Theta\right)=\frac{e^{-En\left(Y,Z,\Theta\right) }}{\mathcal{Z}\left(Y\right)}$. To deduce $P\left(z_{t}|\widetilde{h}_{t},\Theta\right)$, we use Variational Inference (VI) \cite{bishop2006pattern}, a popularly used method in Bayesian inference, which is efficient to handle high-dimensional data. Following VI, we have $P\left(z_{t}|\widetilde{h}_{t},\Theta\right)$ with $Q\left(z_{t}|\widetilde{h}_{t},\Theta\right)$. Then, we minimise the difference between those two distributions using the Kullback–Leibler (KL) divergence metric, which is formulated as follows:

\begin{equation}
\small
\begin{split}
&D_{KL}\left[Q\left(z_{t}|\widetilde{h}_{t},\Theta\right)||P\left(z_{t}|\widetilde{h}_{t},\Theta\right)\right]\\&=\log P\left(\widetilde{h}_{t}|\Theta\right)-E\left[\log P\left(\widetilde{h}_{t}|z_{t},\Theta\right)\right]\\&\quad + D_{KL}\left[Q\left(z_{t}|\widetilde{h}_{t},\Theta\right)||P\left(z_{t}|\Theta\right)\right]
\label{equation:D_KL}
\end{split}
\end{equation}
However, to compute $P\left(\widetilde{h}_{t}|\Theta\right)=\int P\left(\widetilde{h}_{t}|z_{t},\Theta\right)P\left(z_{t}|\Theta\right)dz_{t}$ requires exponential time as it needs to be evaluated over all the configurations of latent variables $z_{t}$. In order to avoid computing $P\left(\widetilde{h}_{t}|\Theta\right)$, we reformulated Eq. (\ref{equation:D_KL}) as an objective function:
\begin{equation}
\small
\begin{split}
ELBO&=\log P\left(\widetilde{h}_{t}|\Theta\right)-D_{KL}\left[Q\left(z_{t}|\widetilde{h}_{t},\Theta\right)||P\left(z_{t}|\widetilde{h}_{t},\Theta\right)\right]\\&=E\left[\log P\left(\widetilde{h}_{t}|z_{t},\Theta\right)\right]\\&\quad-D_{KL}\left[Q\left(z_{t}|\widetilde{h}_{t},\Theta\right)||P\left(z_{t}|\Theta\right)\right]
\label{equation:ELBO}
\end{split}
\end{equation}
where $P\left(\widetilde{h}_{t}|z_{t},\Theta\right)=\prod_{v=1}^{V}p_{\varphi_{t}}\left(\widetilde{h}_{t}^{v}|z_{t}\right)$ with parameters $\varphi_{t}\in\Theta$ under our conditional independence assumption. $p_{\varphi_{t}}$ is a generative network with parameters $\varphi_{t}$ for view $t$. $P\left(z_{t}|\Theta\right)$ is specified as a standard normal distribution $\mathcal{N}\left(0,1\right)$. With the derivation in Supplementary A, we obtain $P\left(z_{t}|\widetilde{h}_{t},\Theta\right)$ below:

\begin{equation}
\small
\begin{split}
P\left(z_{t}|\widetilde{h}_{t},\Theta\right)\approx\frac{\prod_{v=1}^{V}P\left(z_{t}|\widetilde{h}_{t}^{v},\Theta\right)}{\prod_{v=1}^{V-1}P\left(z_{t}|\Theta\right)}
\label{equation:P_z_h}
\end{split}
\end{equation}
That is, $P\left(z_{t}|\widetilde{h}_{t},\Theta\right)$ has the form in which a product of individual posteriors are represented by the priors. We approximate $P\left(z_{t}|\widetilde{h}_{t},\Theta\right)$ with $Q\left(z_{t}|\widetilde{h}_{t},\Theta\right)=\frac{\prod_{v=1}^{V}q_{\phi}\left(z_{t}|\widetilde{h}_{t}^{v}\right)}{\prod_{v=1}^{V-1}P\left(z_{t}|\Theta\right)}$, where  $q_{\phi}\left(z_{t}|\widetilde{h}_{t}^{v}\right)$ is the inference network with parameters $\phi \in \Theta$ in the $v_{th}$ view . For simplicity, each $q_{\phi}\left(z_{t}|\widetilde{h}_{t}^{v}\right)$ is presumably Gaussian with the parameters of mean $\mu_{v}$ and variance $\sigma_{v}$. Then, $Q\left(z_{t}|\widetilde{h}_{t},\Theta\right)$ can be computed as follows:
\begin{equation}
\small
\begin{split}
&Q\left(z_{t}|\widetilde{h}_{t},\Theta\right)\\&=\frac{\prod_{v=1}^{V}q_{\phi}\left(z_{t}|\widetilde{h}_{t}^{v}\right)}{\prod_{v=1}^{V-1}P\left(z_{t}|\Theta\right)}\\&=\frac{\prod_{v=1}^{V}exp\left\{-\frac{1}{2}d\log 2\pi +\gamma_{v}+\mu_{v}^{\top}\sigma_{v}^{-1}z_{t}-\frac{1}{2}z_{t}^{\top}\sigma_{v}^{-1} z_{t}\right\}}{\prod_{v=1}^{V-1}exp\left\{-\frac{1}{2}d\log 2\pi -\frac{1}{2}z_{t}^{\top}z_{t}\right\}}\\&=exp\left\{-\frac{1}{2}d\log 2\pi+\Sigma_{v=1}^{V}\gamma_{v}+\Sigma_{v=1}^{V}\mu_{v}^{\top}\sigma_{v}^{-1}z_{t}\right.\\&\left.\quad -\frac{1}{2}z_{t}^{\top}\left(\Sigma_{v=1}^{V}\sigma_{v}^{-1}-\left(V-1\right)\mathcal{E}_{d}\right)z_{t}\right\}
\label{equation:Q_z_1}
\end{split}
\end{equation}
where $d$ is the dimensionality of latent variables $z_{t}$, $\top$ denotes the transpose operation and $\mathcal{E}_{d}$ is a $d$-by-$d$ identity matrix. $\gamma_{v}$ can be represented as,
\begin{equation}
\small
\begin{split}
&\gamma_{v}=\frac{1}{2}\left(\log |\sigma_{v}^{-1}|-\mu_{v}^{\top}\sigma_{v}\mu_{v}\right)
\label{equation:Q_z_2}
\end{split}
\end{equation}
We observe that $Q\left(z_{t}|\widetilde{h}_{t},\Theta\right)$ is still a Gaussian model with mean $\Gamma=\Lambda \Sigma_{v=1}^{V}\sigma_{v}^{-1}\mu_{v}$ and variance $\Lambda=\left(\Sigma_{v=1}^{V}\sigma_{v}^{-1}-\left(V-1\right)\mathcal{E}_{d}\right)^{-1}$. Hence, the KL divergence between $Q\left(z_{t}|\widetilde{h}_{t},\Theta\right)$ and $P\left(z_{t}|\Theta\right)$ in Eq. (\ref{equation:ELBO}) can be computed as follows:
\begin{equation}
\small
\begin{split}
&D_{KL}\left[Q\left(z_{t}|\widetilde{h}_{t},\Theta\right)||P\left(z_{t}|\Theta\right)\right]\\&=\frac{1}{2}\left(tr\left(\Lambda\right)+\Gamma^{\top}\Gamma-d-\log det\left(\Lambda\right)\right)
\label{equation:ELBO_2}
\end{split}
\end{equation}

\begin{figure*}
\begin{center}
\begin{tabular}{ccc}
\includegraphics[width=4.3cm]{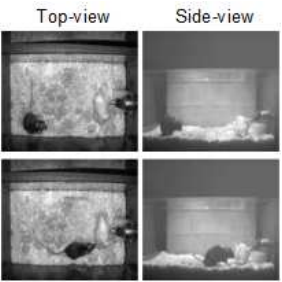}&
\includegraphics[width=7.3cm]{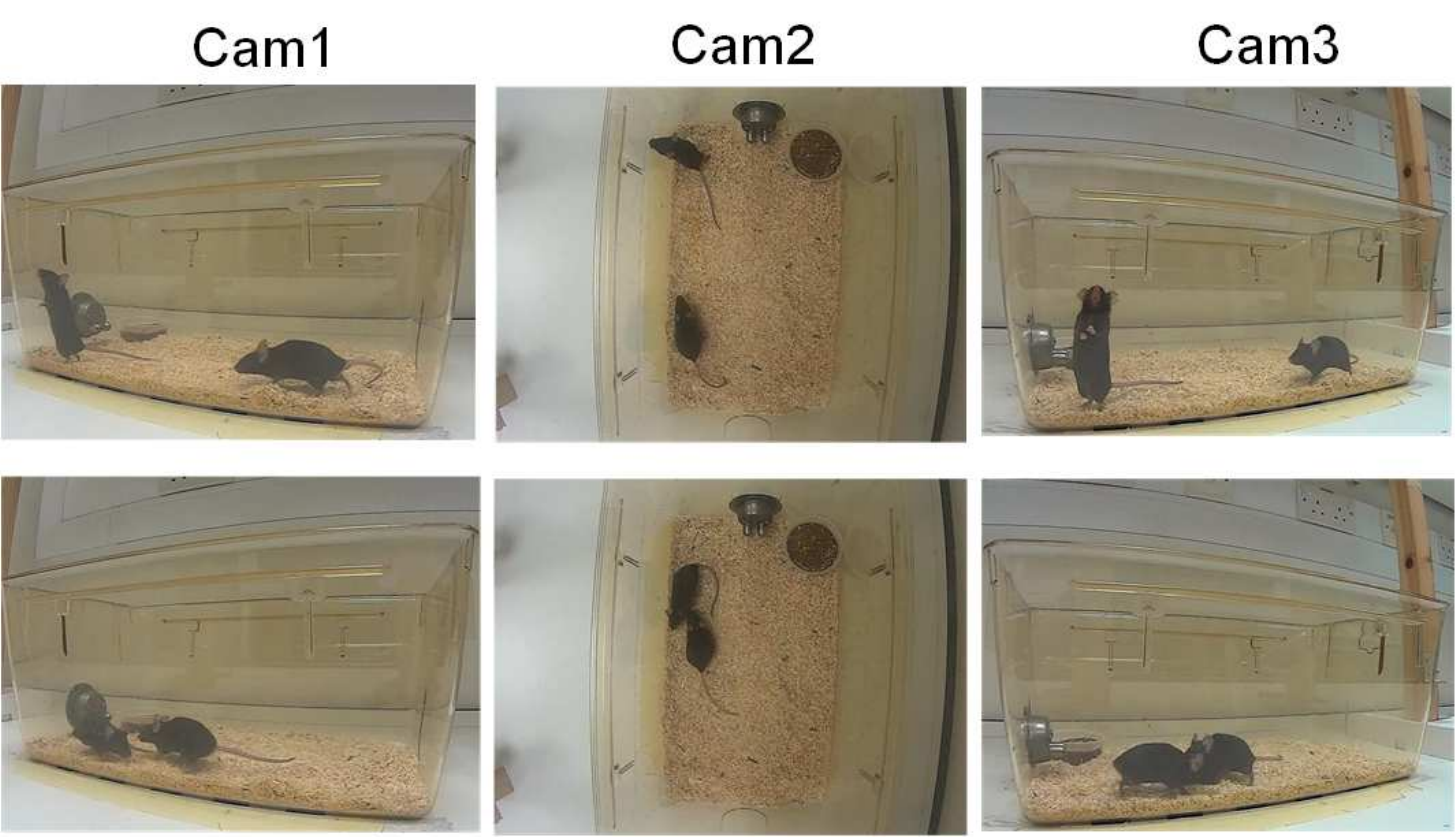}&
\includegraphics[width=6cm]{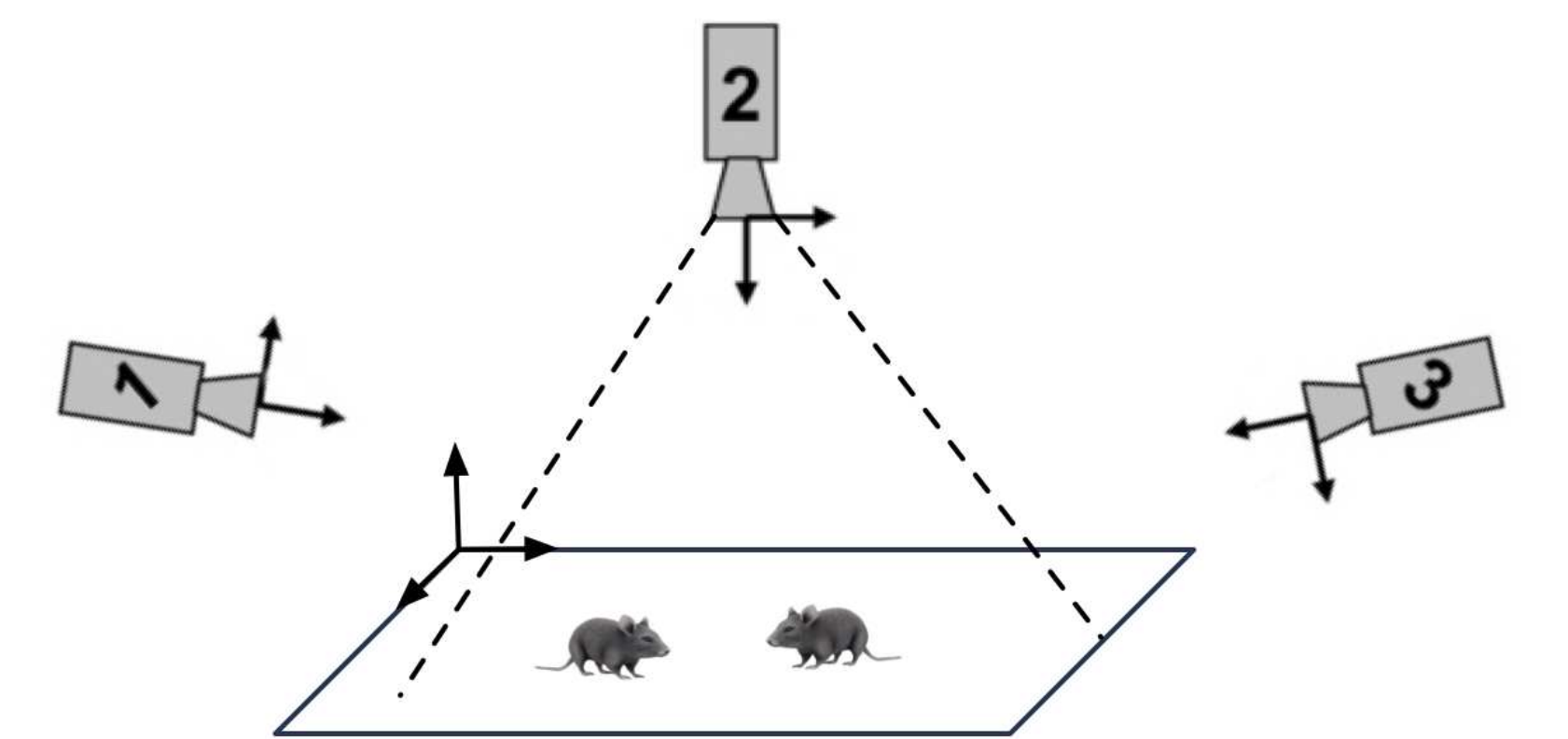}\\\
(a) CRIM13 &(b) our PDMB dataset &(c) camera\\
\end{tabular}
\end{center}
\caption{Snapshots taken from multi-view cameras for the approaching behaviour. The first and second rows in (a) and (b) show the starting and ending frames of the behaviour. (c) illustrates the location of the cameras used in our PDMB dataset. }
\label{fig:snapshots}
\end{figure*}

where, $tr\left(\Lambda\right)$ is a trace function to sum the diagonal elements of matrix $\Lambda$. $dec\left(\Lambda\right)$ is the determinant of matrix $\Lambda$, which can be computed as the product of its diagonals. The whole model can be trained by maximising our ELBO. To address the class imbalance problem during training, we set the sampling rate based on the occurrence frequency of behaviours (as shown in Fig. S1) in the sampling stage of Variational Inference. Although our current model can learn joint representations of the multi-view data, very little information may be missing in each view. As discussed in \cite{burgos2012social}, the top view is suitable to detect behaviors like 'chase' and 'walk away' while the other behaviours, e.g. 'drink' and 'eat', are best recognised from the side view. To utilise such private view information, we adopt a latent attention network to learn the attention weights for both view-shared and view-specific latent variables. For instance, with regards to $V$ views, we can compute $V$ view-specific latent variables $z_{t}\left(\widetilde{h}_{t}^{1}\right),...,z_{t}\left(\widetilde{h}_{t}^{V}\right)$ and $2^{V}-V-1$ view-shared latent variables $z_{t}\left(\widetilde{h}_{t}^{1},\widetilde{h}_{t}^{2}\right),...,z_{t}\left(\widetilde{h}_{t}^{1},...,\widetilde{h}_{t}^{V}\right)$. Hence, the expectation of $z_{t}$ in Eq. (\ref{equation:approximation_Z}) can be calculated as:

\begin{equation}
\small
\begin{split}
E\left[z_{t}\right]=\alpha_{1,n} E\left[z_{t}\left(\widetilde{h}_{t}^{1}\right)\right]+...+\alpha_{2^{V}-1,n}E\left[z_{t}\left(\widetilde{h}_{t}^{1},...,\widetilde{h}_{t}^{V}\right)\right]
\label{equation:average_Z}
\end{split}
\end{equation}
where $\alpha_{i,n}$ is a score assigned to each latent variable based on its relevance to the behavioural label $b_{n}\in \mathcal{Y}$. We calculate $\alpha_{i,n}$ as follows:

\begin{equation}
\small
\begin{split}
\alpha_{i,n}=\frac{exp\left(r_{i,n}\right)}{\Sigma_{i=1}^{j}exp\left(z_{t,i}\right)}
\label{equation:average_Z}
\end{split}
\end{equation}
where $r_{i,n}=Em(b_{n})U_{n}r_{i,n}$ is the attention score measuring the relationship between the latent variable $z_{t,i}$ and the behavioural label $b_{n}$. $Em$ is a word embedding function which is widely used on natural language processing. The weight matrix $U_{n} \in \Theta$ is the parameter to be learned.

To calculate $E\left[h_{t}^{v}\right]=\sum_{h_{t}^{v}}h_{t}^{v}P\left(h_{t}^{v}|x_{t}^{v},\Theta\right)$ in Eq. (\ref{equation:average_h}), we adopt the classical LSTM. Then, we can have:

\begin{equation}
\small
\begin{split}
P\left(h_{t}^{v}|x_{v}\right)&=\frac{e^{-En\left(H,X,\Theta\right) }}{\mathcal{Z}\left(H\right)}\\&=\frac{1}{\mathcal{Z}\left(h_{t}^{v}\right)}e^{-\mathcal{E}\left(h_{t}^{v},x^{v}\right)-\mathcal{E}\left(h_{t-1}^{v},h_{t}^{v}\right)}\\
\label{equation:average_Z}
\end{split}
\end{equation}
where $\mathcal{E}\left(h_{t}^{v},x^{v}\right)=W_{o}x_{v}^{t}$ and $\mathcal{E}\left(h_{t-1}^{v},h_{t}^{v}\right)=U_{o}h_{t-1}+b_{o}$ are defined in traditional Recurrent Neural Network (RNN), while LSTM has an extra state called cell which is protected and controlled by the three gates. Hence, $E\left[h_{t}^{v}\right]$ can be calculated below:
\begin{equation}
\small
\begin{split}
E\left[h_{t}^{v}\right]&=c_{\Pi}^{t}\Sigma_{h_{t}^{v}}h_{t}^{v}P\left(h_{t}^{v}|x_{v}\right)\\&=c_{\Pi}^{t}sigm(W_{o}x_{v}^{t}+U_{o}h_{t-1}+b_{o})
\label{equation:LSTM}
\end{split}
\end{equation}
where $c_{\Pi}^{t}$ with parameter $\Pi$ has the same definition as LSTM, and both $\Pi$, $W_{o}$, $U_{o}$ and $b_{o}$ are parameters to be learned.
The implementation of MV-LADDM is available from \href{https://github.com/ZhehengJiang/MV-LADDM.git}{https://github.com/ZhehengJiang/MV-LADDM.git}. \\

\section{Experimental work}
\subsection{Video database}
\subsubsection{CRIM13 dataset}
In this section, we firstly give an overview  of a publicly available multi-view mouse social behaviour dataset: the Caltech Resident-Intruder Mouse (CRIM13) dataset \cite{burgos2012social}. This dataset was used to study neurophysiological mechanisms in the mouse brain. It consists of 237*2 videos that was recorded using synchronised top- and side-view cameras with the resolution of 640*480 pixels and the frame rate of 25Hz. Each video lasts around 10 minutes and was annotated frame by frame. There are 12+1 different mutually exclusive behaviour categories, i.e. 12 behaviors and one otherwise unspecified behaviour for the description of mouse behaviours. Fig. \ref{fig:snapshots}(a) shows video frames for the approaching behavior in both top and side views. The occurrence probabilities of behaviours are expressed as percentages in Fig. S1(a). The behaviours in CRIM13  are highly imbalanced. Except for `other' (56.0\%) behaviours, the most occurring behaviour is `sniff' (13.9\%), and the least occurring behaviours are `circle' and `drink' (only 0.4\%). We also show the occurrence frequency of the neighbouring labels in Fig. S2(a). From the figure, we can observe there is a strong correlation between different behaviours. For example, it is very unlikely to have a `circle’, `drink’, `eat’ or `clean’ behaviour immediately after an `approach’ behaviour, as the occurrence frequency of the former behaviours is zero. This has motivated us to model such label correlation in our graphic model.

\subsubsection{PDMB dataset}
In this paper, we introduce a new dataset, which was collected in collaboration with the biologists of Queen’s University Belfast of United Kingdom, for a study on motion recordings of mice with Parkinson’s disease (PD). The neurotoxin 1-methyl-4-phenyl-1,2,3,6-tetrahydropyridine (MPTP) is used as a model of PD, which has become an invaluable aid to produce experimental parkinsonism since its discovery in 1983 \cite{sedelis2001behavioral,
jackson2002blockade,jackson2007protocol,handa2019bone}. Six C57bl/6 female mice received treatment of MPTP while other six wild-type female mice are used as controls. All the mice used throughout this study were housed (3 mice of the same type per cage) in a controlled environment with the constant temperature of ($27^{\circ}C$) and light condition (long fluorescent lamp of 40W), and under constant climatic conditions with free access to food and water (placed on the corner of the cage). All experimental procedures were performed in accordance with the Guidance on the Operation of the Animals (Scientific Procedures) Act, 1986 (UK) and approved by the Queen’s University Belfast Animal Welfare and Ethical Review Body.

The proposed dataset consists of 12*3 annotated videos (6 videos for MPTP treated mice and 6 videos for control mice) recorded by using three synchronised Sony Action cameras (HDR-AS15) (one top-view and two side-view) with the frame rate of 30 fps and video resolution of 640 by 480 pixels. Fig. \ref{fig:snapshots}(b) and (c) show video frames for the approaching behavior in three views and the locations of our cameras. We follow the behaviour definition of CRIM13 \cite{burgos2012social} (see Table \ref{tab:CRIM13_description}) to annotate all the videos in the PDMB dataset. All the videos (216,000*3 frames in total) contain 8+1 behaviours of two freely behaving mice and each video lasts around 10 minutes. Activity occurrences and the occurrences of neighbouring activities are shown in Fig. S1(b) and Fig. S2(b) respectively.

\subsection{View-specific feature representation}
To extract view-specific features, sliding windows are centered at each frame, wherein all types of view-specific features are sought. The method of extracting view-specific features is adapted from the previous works for single-view mouse behaviour recognition\cite{jiang2018context,jiang2017behavior}. We adopt spatio-temporal and trajectory-based motion features as both of them result in satisfactory performance \cite{jiang2018context}. More technical details can be found in the Proposed Methods section.

\begin{figure*}
\begin{center}
\begin{tabular}{ccc}
\includegraphics[width=5.8cm]{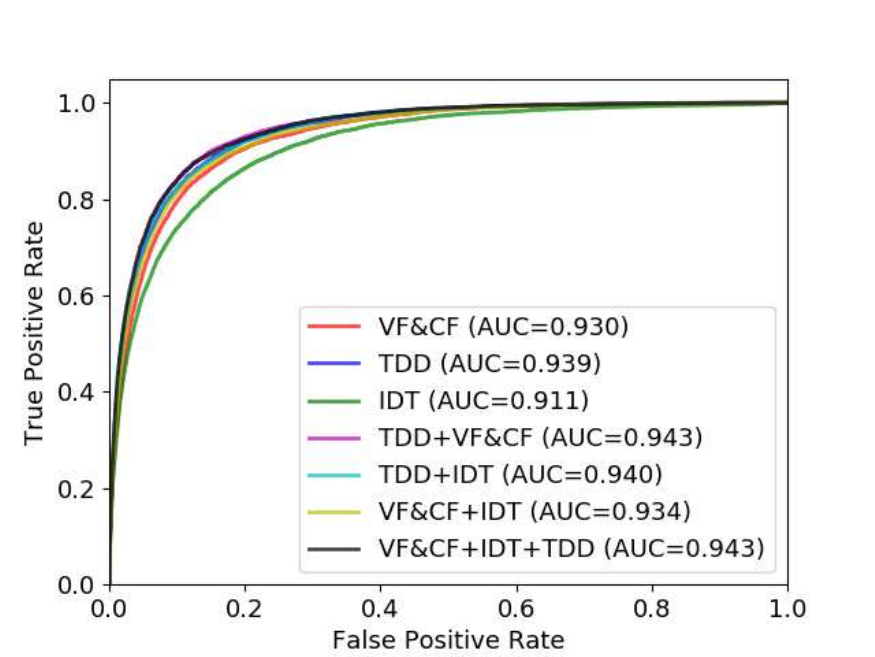}&
\includegraphics[width=5.8cm]{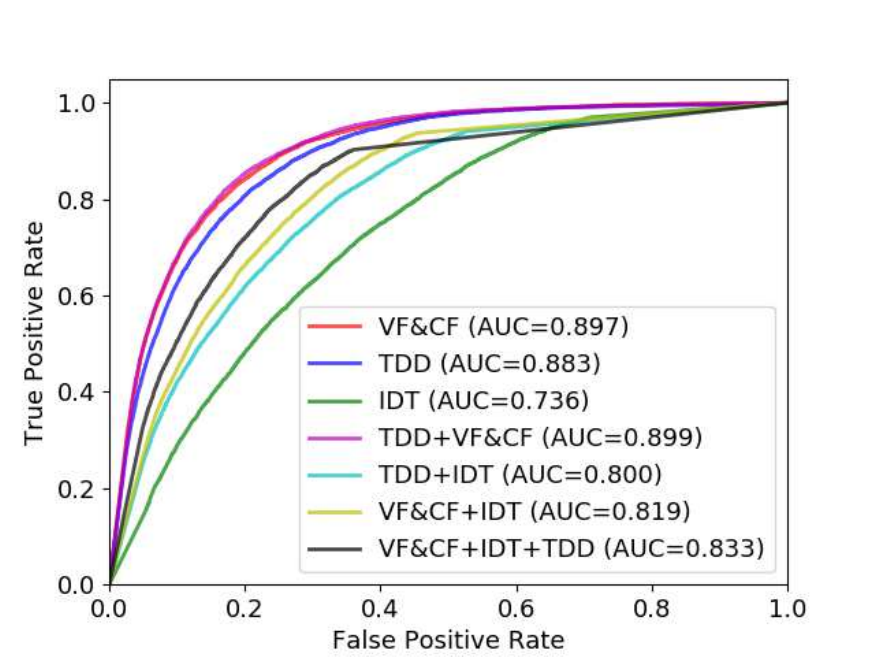}&
\includegraphics[width=5.8cm]{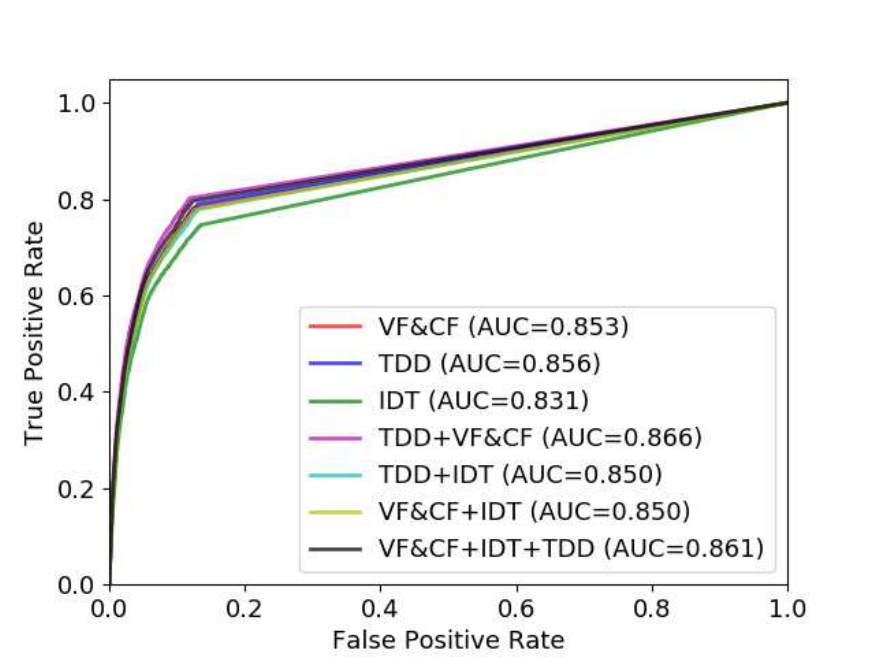}\\
(a) LR&(b) BNB&(c) KNN\\
\includegraphics[width=5.8cm]{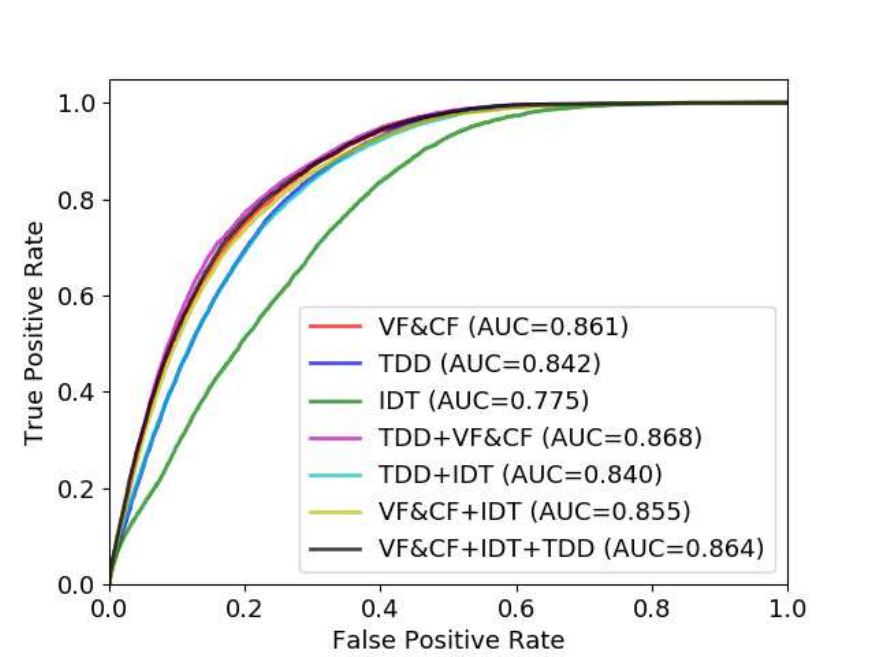}&
\includegraphics[width=5.8cm]{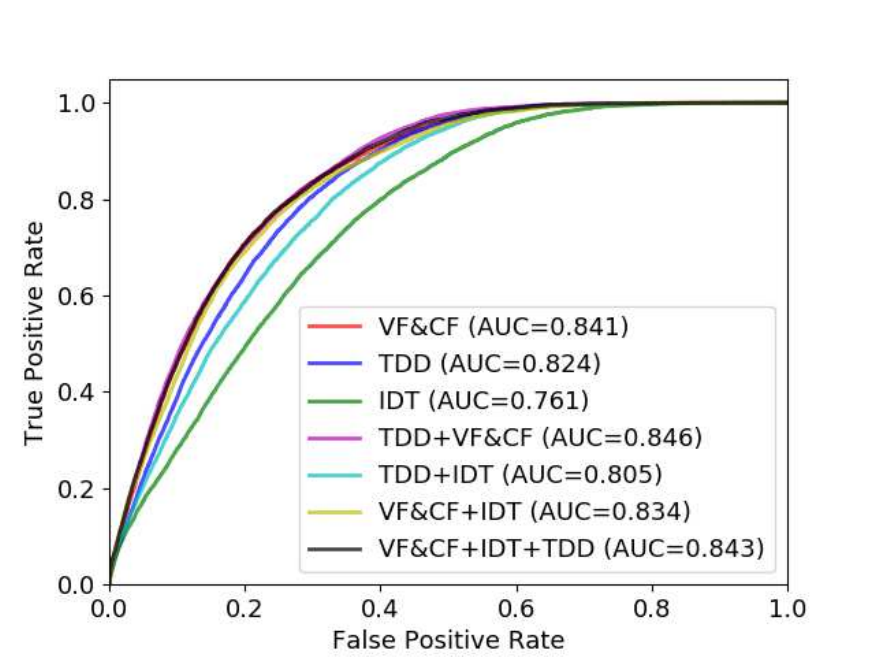}&
\includegraphics[width=5.8cm]{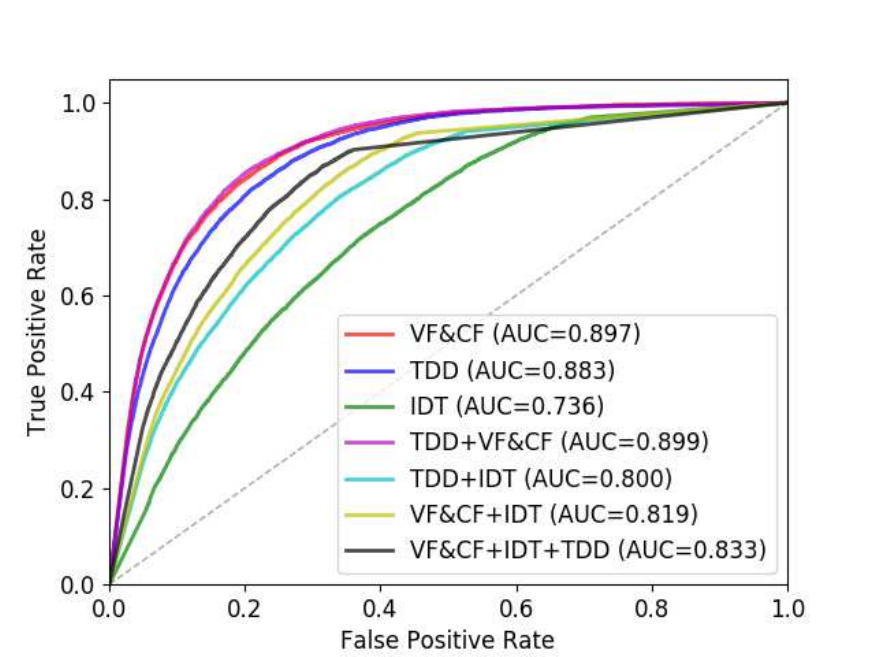}\\
(d) AdaB&(e) RF&(f) SVM
\end{tabular}
\end{center}
\caption{Receiver Operating Characteristic (ROC) curves of the classification outcome for the CRIM13 dataset. These classifiers include (a) Logistic Regression, (b) Bernoulli naive Bayes, (c) 5-nearest neighbours, (d) AdaBoost with the base estimator of Random forest, (e) Random forest, and (f) Support Vector Machine with a linear kernel.}
\label{fig:diff_methods}
\end{figure*}

\begin{table*}
\begin{center}
\caption{Performance (accuracy) of using different features on the CRIM13 dataset. Accuracy figures are reported as the averaging one across all the behaviours where the chance level is 7.69\% for an thirteen-class classification problem. We observe that the combination of TDD and VF\&CF  is able to achieve relatively high accuracy with individual classifiers. Particularly, for BNB, KNN and SVM,  the combined features result in higher accuracy than individual uses of the features. In comparison, features combined with IDT leads to worse system performance, and the performance of the other features is significantly worse than that of using TDD and VF\&CF features together. }
\label{tab:specific_behaviours}
\begin{tabular}{|p{2.6cm}|p{2.4cm}|p{1.3cm}|p{1.3cm}|p{1.3cm}|p{1.3cm}|p{1.3cm}|p{1.3cm}|p{1.3cm}|}
\hline
\multicolumn{2}{|c|}{Feature extraction method} & LR & BNB & KNN & AdaB & RF & SVM & Average\\
\hline\hline
\multirow{2}{2.4cm}{\textit{Trajectory-based motion features}}&IDT\cite{wang2013action}  & 29.7\% & 28.0\% & 22.7\% & 24.2\% & 34.3\% & 22.4\% & 26.9\%\\
&TDD\cite{wang2015action} & \textbf{33.6\%} & 40.5\% & 30.2\% & 35.5\% & 36.3\% & 26.7\% & 33.8\%\\
\hline
\multirow{4}{2.4cm}{\textit{Spatio-temporal features}}&VF\&CF\cite{jiang2018context} & 32.4\% & \textbf{41.3\%} & 27.4\% & \textbf{40.3\%} & \textbf{39.8\%} & 25.2\% & \textbf{34.4\%}\\
&Harris3D\cite{burgos2012social} & $\backslash$ & $\backslash$ & $\backslash$ & 20.9\% & $\backslash$ & $\backslash$ & 20.9\%\\
&Cuboids\cite{burgos2012social} & $\backslash$ & $\backslash$ & $\backslash$ & 24.6\% & $\backslash$ & $\backslash$ & 24.6\%\\
&LTP\cite{burgos2012social} & $\backslash$ & $\backslash$ & $\backslash$ & 22.2\% & $\backslash$ & $\backslash$ & 22.2\%\\
\hline
\multirow{4}*{\textit{Combined features}}&TDD+VF\&CF & \textbf{33.5\%} & \textbf{42.2\%} & \textbf{35.5\%} & \textbf{39.5\%} & \textbf{38.9\%} & \textbf{27.5\%} & \textbf{36.2\%}\\
&TDD+IDT & 32.7\% & 33.1\% & 30.4\% & 31.5\% & 29.4\% & 26.5\% & 30.6\%\\
&VF\&CF+IDT & 30.9\% & 33.2\% & 28.2\% & 35.2\% & 35.0\% & 24.9\% & 31.2\%\\
&VF\&CF+IDT+TDD & 32.7\% & 34.7\% & \textbf{33.0\%} & 37.5\%& 38.3\% & \textbf{27.0\%} & 33.9\%\\
\hline
\textit{Deep learned features} & Two-stream \cite{simonyan2014two} & 26.1\% & 25.7\% & 19.3\% & 21.3\%& 30.3\% & 20.2\% & 23.8\%\\
\hline
\end{tabular}
\end{center}
\end{table*}

To evaluate the contribution of these features towards the recognition of mouse behaviours, as an example, we wish to examine different classifiers over sliding windows on the top-view videos. These approaches neither rely on the multi-view feature fusion nor the temporal context of mouse behaviours, corresponding to the view-specific features. To this end, we collect a subset of the CRIM13 dataset which was also used in \cite{burgos2012social} for analysing their feature extraction method. This small validation dataset includes 20 top-view videos randomly chosen from the whole dataset and is evenly divided to training and testing datasets. We assess some of the most widely used trajectory-based motion features, spatio-temporal features and their combinations. In the approaches based on trajectory-based motion features, we use the established Improved Dense Trajectory (IDT) technique, which densely samples image points and tracks them using optical flows. In the evaluation, we deploy the default trajectory length of 15 frames. For each trajectory, we compute Histograms of Oriented Gradients (HOG), Histograms of Optical Flow (HOF) and Motion Boundary Histograms (MBH) descriptors proposed in \cite{wang2013action}. The final dimensions of the descriptors are, 96 for HOG, 108 for HOF and 192 for MBH. Another Trajectory-based motion feature extraction approach in our assessment is the trajectory-pooled deep convolutional descriptor (TDD) \cite{wang2015action}. The goal of TDD is to combine the benefits of both trajectory-based and deep-learned features. This local trajectory-aligned descriptor is computed from the spatial and temporal nets. Following their default settings, we use the descriptors from conv4 and conv5 layers for the spatial nets, and conv3 and con4 layers for the temporal nets. These networks are pre-trained on ImageNet \cite{simonyan2014two} and fine-tuned on the UCF-101 \cite{burgos2012social} dataset. Finally, we concatenate these descriptors and reduce the dimensionality of the vector using Principal Component Analysis (PCA) (256 components are kept as default). We also evaluate the performance of Two-Stream Convolutional Networks \cite{simonyan2014two}, which is a popular deep learning model for human action recognition. We fuse the outputs of the last fully-connected layers of spatial and temporal nets and obtain 4096 dimensional feature vectors. 

\begin{table*}
\begin{center}
\caption{View-shared behaviour recognition results of various approaches on the CRIM13 dataset. In Deep Canonical Correlation Analysis (DCCA) \cite{andrew2013deep}, we change the node number of its output layer from 50 to 150 in our experiment. In Kernel Canonical Correlation Analysis (KCCA), we adopt linear \cite{hotelling1992relations}, Gaussian and polynomial kernels \cite{lai2000kernel} for comparison. It shows that our approach achieves the best recognition performance for 11 out of 12 behaviours.}
\label{tab:view_shared_performance}
\begin{tabular}{|p{1.4cm}|p{1.4cm}|p{1.4cm}|p{1.4cm}|p{1.4cm}|p{1.4cm}|p{1.4cm}|p{2.5cm}|}
\hline
\multirow{2}{*}{Behaviour} & \multicolumn{3}{|c|}{DCCA} & \multicolumn{3}{|c|}{KCCA} &\multirow{2}{*}{\textbf{Ours (View-shared)}} \\
\cline{2-7}
&50 & 100&150 & Linear &Gaussian &Polynomial &\\
\hline\hline
approach & 18.0\% & 45.7\% & 52.9\% & 54.0\% & 54.4\% & 54.3\%& \textbf{58.1\%}\\
attack & 48.9\% & 61.5\% & 52.2\% & 61.1\% & 65.1\% & 60.5\% & \textbf{67.2\%}\\
copulation  & 22.0\% & 48.2\% & 37.7\% & 60.6\% & 59.2\% & 59.0\% & \textbf{68.3\%}\\
chase & 16.8\% & 31.4\% & 31.8\% & 29.6\% & 31.4\% & 29.1\% & \textbf{38.2\%}\\
circle & 0\% & 5.1\% & 8.9\% & 8.5\% & 8.5\% & 9.3\% & \textbf{34.3\%}\\
drink & 26.3\% & 63.8\% & 67.5\% & 45.0\% & 38.8\% & 43.8\% &\textbf{87.5\%}\\
eat & 55.0\% & 78.3\% & 94.8\% & \textbf{99.0\%}& 98.7\% & 98.4\% &40.8\%\\
clean & 51.6\% & 75.0\% & 74.0\% & 78.2\%& 80.6\% & 80.0\% &\textbf{81.7\%}\\
human & 62.3\% & 89.1\% & 92.0\% & 94.9\% & 93.1\% & 88.0\% &\textbf{98.3\%}\\
sniff & 28.2\% & 39.9\% & 33.2\% & 50.6\% & 52.2\% & 50.1\% &\textbf{57.1\%}\\
up & 39.3\% & 77.8\% & 78.9\% & 66.6\% & 67.7\% & 67.5\% &\textbf{80.2\%}\\
walk away & 15.5\% & 39.9\% & 47.0\% & 53.7\% & 52.9\% & 50.4\% &\textbf{57.7\%}\\
other & 86.9\% & 92.7\% & \textbf{93.9\%} & 93.4\% & 93.7\% & 92.5\% &49.4\%\\
{\bf Average} & 36.2\% & 57.6\% & 58.8\% & 60.9\% & 61.3\% & 60.2\% &63.0\%\\
\hline
\end{tabular}
\end{center}
\end{table*}

A large number of papers published so far have shown the promising performance of the above approaches on human action datasets, but very few papers are related to the exploration of mouse behaviours. Popularly used spatio-temporal feature extraction approaches include VF\&CF\cite{jiang2018context}, Harris3D\cite{burgos2012social}, Cuboids\cite{burgos2012social}, and LTP\cite{burgos2012social}. In our experiments, all the parameters used in these approaches have been set to their original configurations which give the best results in behaviour recognition of mice \cite{jiang2018context,burgos2012social}. We incorporate these features with individual classifiers and illustrate the classification results in Table \ref{tab:specific_behaviours}, where the classifiers include Logistic Regression (LR), Bernoulli naive Bayes (BNB), 5-nearest neighbours (KNN), Random forest (RF), AdaBoost (AdaB) with the base estimator of RF and Support Vector Machine (SVM) with a linear kernel. We also report their average accuracy in the table. The highlighted figures in the table demonstrate that the use of TDD, VF\&CF and their combination usually result in the best classification accuracy. In particular, for BNB, KNN and SVM,  the combined features are able to achieve better accuracy than the individual use of them. The effectiveness of the other schemes, including Two-stream Convolutional Networks and IDT, are significantly lower than that of TDD and VF\&CF. Note that VF\&CF can achieve 15.7\% better than Cuboids that has been reported to achieve the best performance \cite{burgos2012social}. It is observed that the features combined with IDT deteriorate the system performance. In fact, complementary features perform much better than casual feature combination with regards to system accuracy. Receiver Operating Characteristic (ROC) curves of individual classifiers with different feature combinations and their area under the curve (AUC) are shown in Fig. \ref{fig:diff_methods}. We also witness that the combination of TDD and VF\&CF have the highest AUC, the best performance in each classifier.

%-------------------------------------------------------------------------
\subsection{Social Behaviour Recognition}
In our system, for the efficiency purpose, all the view-specific features are computed from a small sliding window in the video (length = 40 frames), which are centered at each frame. Our system aims at assigning every sliding window to one of the pre-defined behaviour categories. For this challenging task, the temporal and view contexts of each specific behaviour are fully utilised in our system. To do so, we propose a novel Multi-view Latent-Attention Dynamic Discriminative Model that includes (1) the modelling of the temporal relationship of image frames for each segment, (2) the modelling of the relationship between views, and (3) the modelling of the correlations between the labels in neighbouring regions. Details about the system implementation are provided in the Proposed Methods section. For efficiency and simplification, we divide the experiments in this section into two parts: View-Shared and View-Attention Behaviour Recognition.

Traditionally, classification accuracy is defined as the percentage of the samples that are correctly labelled against the number of the overall samples. While being a valid measure, this metric cannot disclose the characteristics of the datasets that have a severe imbalanced classification problem. To better measure the system performance, we here use the averaging recognition rate per behaviour.

\subsubsection{View-Shared Behaviour Recognition}
In this experiment, we leave the view-specific features and only use the learned view-shared features. For the fair comparison, we adopt the same classifier (i.e. linear SVM) and compare its recognition results with those of canonical correlation analysis (CCA) \cite{hotelling1992relations}, kernel CCA (KCCA) \cite{lai2000kernel} and deep CCAs \cite{andrew2013deep}, resulting in Table \ref{tab:view_shared_performance}. It is worth pointing out that CCA is a way of measuring the linear relationship between two views in the projected space. KCCA is the extension of the standard CCA, where explicit mapping to the feature space can be avoided and the correlation can be performed in the feature space by replacing the scalar products with the kernel function in the input space. We adopt Gaussian and polynomial kernels for the comparison in this study. DCCA \cite{andrew2013deep} is introduced to address the scalability issue using deep learning and we vary the node number of its output layer from 50 to 150 in our experiment for deeper exploration. As shown in Table \ref{tab:view_shared_performance}, our approach achieves the best recognition rate for 11 out of 12 behaviours, significantly better than the other state of the art approaches. It also demonstrates the effectiveness of our learned features. Moreover, using Variational Inference (more details can be found in the Proposed Methods section), our model can effectively handle the overfitting problem with the strength of dealing with the imbalanced data.

\begin{table*}
\begin{center}
\caption{Behaviour recognition results of various approaches for the CRIM13 dataset.}
\label{tab:attention}
\begin{tabular}{|p{1.2cm}|p{0.8cm}|p{1.1cm}|p{0.8cm}|p{1cm}|p{1cm}|p{1cm}|p{0.8cm}|p{0.8cm}|p{1.5cm}|p{1.6cm}|p{1.4cm}|}
\hline
\multirow{2}{1.2cm}{Behaviour} & \multirow{2}{0.8cm}{PBMV} & \multirow{2}{1.3cm}{KCCA (Gaussian)} & \multirow{2}{1.3cm}{DCCA} & \multirow{2}{1.3cm}{BILSTM} & \multirow{2}{1.3cm}{DCLSTM} & \multirow{2}{1.2cm}{Burgos-Artizzu et al. }&\multirow{2}{1cm}{LDCRF}&\multirow{2}{1cm}{MV-LDCRF}&
\multirow{2}{1.5cm}{Ours (View-shared)}&
\multicolumn{2}{|c|}{Ours (View-attention)}\\\cline{11-12}
& & & & & & & & & &Ours (without label correlation)& Ours (with label correlation)\\
\hline\hline
approach & 7.2\% & 54.4\% & 52.9\% & 33.8\% & 28.1\% & \textbf{75.0\%} &34.9\%& 52.8\% &58.1\% & 51.6\% &51.8\%\\
attack & 86.3\% & 65.1\% & 52.2\% & \textbf{96.1\%} & 96.0\% & 59.0\% &70.2\%&68.4\%& 67.2\% &71.8\% &73.0\%\\
copulation  & 21.2\% & 59.2\% & 37.7\% & 11.0\% & 13.3\% & 62.0\% &47.5\%& 59.7\%& 68.3\% &70.4\%& \textbf{71.2\%}\\
chase & 64.8\% & 31.4\% & 31.8\% & 79.0\% & \textbf{86.8\%} & 70.0\% &36.1\%& 41.9\%& 38.2\% & 53.2\% & 62.3\%\\
circle & 0.7\% & 8.5\% & 8.9\% & 47.8\% & 43.1\% & \textbf{68.0\%} &41.4\%&57.6\%& 34.3\% & 58.9\% &67.8\%\\
drink & 97.5\% & 45.0\% & 67.5\% & 94.2\% & 89.2\% & 49.0\% &36.3\%& 78.2\%&87.5\% & 93.8\% & \textbf{97.5\%}\\
eat & 1.2\% & \textbf{98.7\%} & 94.8\% & 95.1\%& 93.8\% & 53.0\% &11.0\%&46.4\%& 40.8\%& 43.7\% &61.8\%\\
clean & 60.7\% & 80.6\% & 74.0\% & 62.1\%& 61.1\% & 47.0\% &60.4\%&78.5\%& 81.7\% & 69.4\% &\textbf{82.4\%}\\
human & 32.1\% & 93.1\% & 92.0\% & 47.5\% & 41.3\% & 96.0\% &38.3\%& 66.4\%&98.3\% & \textbf{99.4\%} &98.3\%\\
sniff & 28.3\% & 52.2\% & 33.2\% & 36.0\% & 38.8\% & 44.0\% &49.7\%& 58.9\%& 57.1\% & 61.6\% &\textbf{66.2\%}\\
up & 94.8\% & 67.7\% & 78.9\% & \textbf{94.2\%} & 93.0\% & 62.0\% &64.6\%&71.6\%& 80.2\% & 79.2\% &79.2\%\\
walk away & 13.2\% & 52.9\% & 47.0\% & 19.3\% & 19.2\% & 56.4\% &29.8\%&54.7\%& 57.7\% & 56.3\% &\textbf{58.5\%}\\
other & 94.6\% & 93.7\% & 93.9\% & 94.7\% & \textbf{95.3\%} & 53.0\% &69.4\%&68.2\%& 49.4\% & 45.0\% &61.5\%\\
{\bf Average} & 46.4\% & 61.3\% & 58.8\% & 62.4\% & 62.2\% & 62.6\% &45.4\%&61.7\%& 63.0\% & 65.7\% & \textbf{71.7\%}\\
\hline
\end{tabular}
\end{center}
\end{table*}

\subsubsection{View-Attention Behaviour Recognition}
This experiment is prepared with both the view-specific and view-shared features, where the former captures unique dynamics of each view whilst the latter encodes the interaction between the views. In our proposed model, attention scores are automatically learned to measure the contributions of each view-specific and view-shared feature in the recognition of mouse behaviours. Our view-attention behaviour recognition approach is compared against the existing approaches such as  \cite{poria2018meld,hotelling1992relations,lai2000kernel,andrew2013deep,goyal2019multiview,zhao2018dual}. The PBMVboost \cite{goyal2019multiview} is a two-level multi-view learning approach, which learns the distribution over view-specific classifiers or views in one single step by a boosting approach. The number of the iterations used in PBMVboost is set to 100 with a tree depth 13 (class number), experimentally. CCA, KCCA and DCCA can report the correlation over the representations from different views, but how to utilise the view-specific information is not addressed in these approaches. BcLSTM\cite{poria2018meld} and DCLSTM \cite{zhao2018dual} are two Long Short-Term Memory (LSTM) based approaches with specific hyperparameters set to the optimum values (epochs: 100, batch size: 50, and learning rate: 0.001). We also compare our approach to the baseline graphical model LDCRF \cite{morency2007latent} and its multi-view counterpart, i.e. MV-LDCRF \cite{song2012multi}. For LDCRF, the final class scores are obtained by averaging the scores of different views. All the parameters of LDCRF and MV-LDCRF are set to the default. To reduce their computational cost, the dimensions of all the features are reduced to 1000 using PCA. The importance of modelling the correlations between the neighbouring labels in our approach is also evaluated.

Table \ref{tab:attention} depicts that our approach with modelling label correlation achieves the highest averaging accuracy, i.e. 71.7\%. Our view-attention approach outperforms the view-shared approach, suggesting the effectiveness of adding the attention model into the framework. Without using view-specific features, only using the shared features cannot make the data discriminative enough for satisfactory classification, especially in the cases where features are not shared across different views. The methods, e.g. \cite{goyal2019multiview,poria2018meld,zhao2018dual}, also exploit view-specific features. They treat the features across views equally and thus cannot properly value the importance of the features collected from different views. In Table \ref{tab:attention}, we observe that BILSTM and DCLSTM have poor performance (accuracy is lower than 20\%) in the recognition of `copulation' and `walk away'.

\begin{figure*}
\begin{center}
\includegraphics[width=16cm]{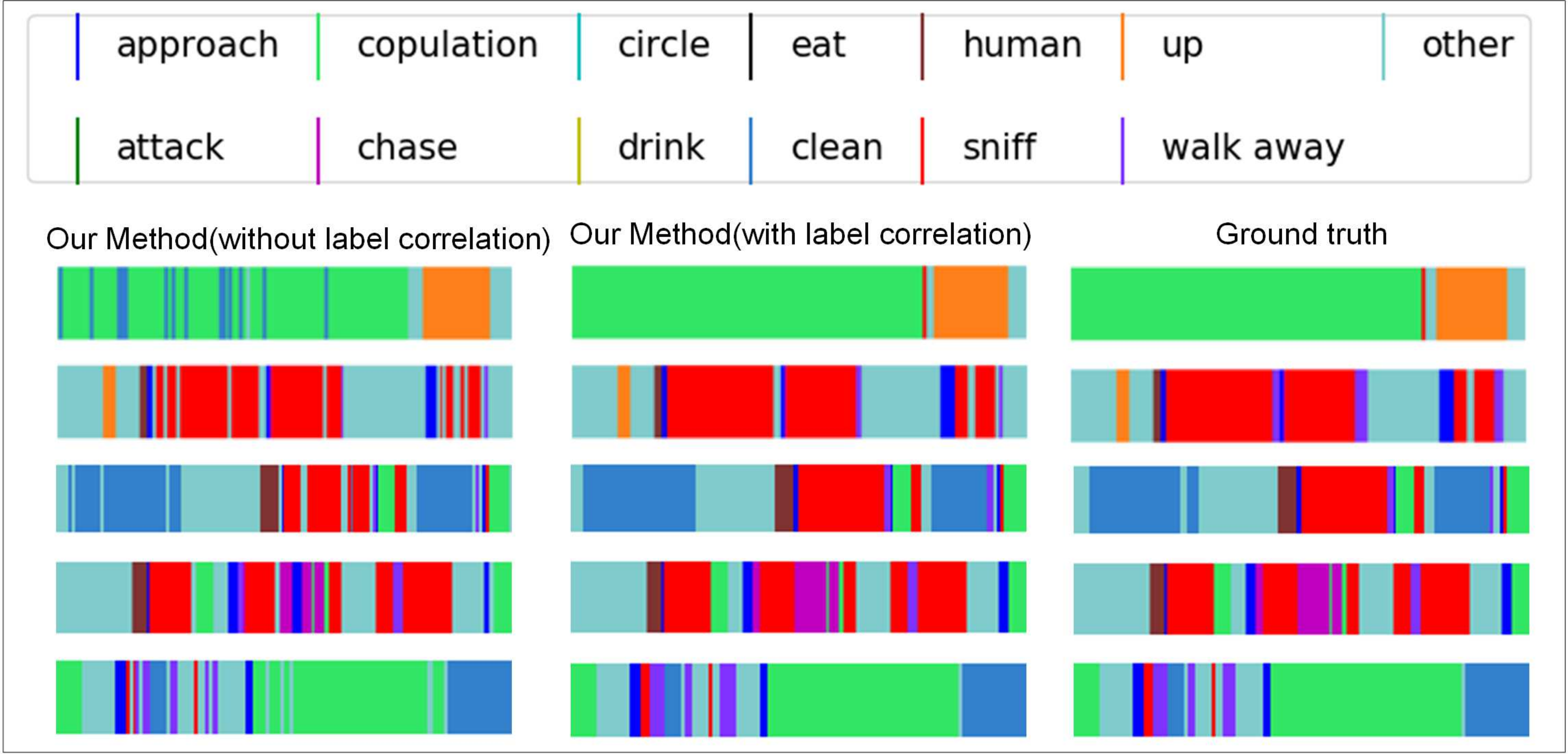}
\end{center}
\caption{Comparison of the chronograms of the ground-truth and our method for the test video. The necessity of modelling label correlation can be observed in this figure. The average agreement rate between the labeling results by our approach and the expert annotators can be found in Fig. S3.}
\label{fig:chronogram}
\end{figure*}

The importance of the modelling label correlation is clearly demonstrated in Figs. S3 and \ref{fig:chronogram}. Our two approaches achieve superior performance over all the other approaches, demonstrating the benefit of modelling label correlation and attention modelling in this experiment. Fig. S4 shows the average agreement rates of our approaches over 2-, 4- and 6-minute intervals. For statistical analysis, two-sample t-test and paired t-test are performed under the assumption of Gaussian errors. Wilcoxon signed-rank tests are also used to examine this assumption. All the testing results suggest that our method with label correlation significantly improves the average agreement rate ($p1<p2<p3<0.05$). Furthermore,  We do not see any significant difference in the mean average agreement rates over various intervals, shown in Fig. S4(a), (b) and (c), suggesting the performance of our approaches does not go down over time. In addition, our approach is robust against viewpoint variations and can achieve satisfactory performance in multi-view recognition. The time cost of training and testing different systems is reported in Table \ref{tab:time_training_testing}. All the algorithms are implemented on a PC with a 3.6-GHz Intel Core i7 processor and a NVIDIA RTX 2080Ti GPU. Since MV-LDCRF approach only provides the code with CPU implementation, we follow its default setting and report its testing time on our CPU. From Table V, we can see our approach achieves competitive speed, compared against the other state-of-the-art approaches. 

\begin{table}
\begin{center}
\caption{Time of training and testing different systems.}
\label{tab:time_training_testing}

\begin{tabular}{|p{1.3cm}|p{0.6cm}|p{0.6cm}|p{0.9cm}|p{1cm}|p{0.9cm}|p{0.5cm}|}
\hline
 & PBMV & DCCA &  BILSTM & DCLSTM & MV-LDCRF & our\\
\hline\hline
Training (hour) & 26.1 & 19.6 & 15.2 & 16.1 & 63.4 & 18.5 \\
\hline
Testing (min) & 2.1 & 1.4 & 0.9 & 1.1 & 10.2 & 1.2 \\
\hline
\end{tabular}
\end{center}
\end{table}

\begin{figure}
\begin{center}
\includegraphics[width=7.5cm]{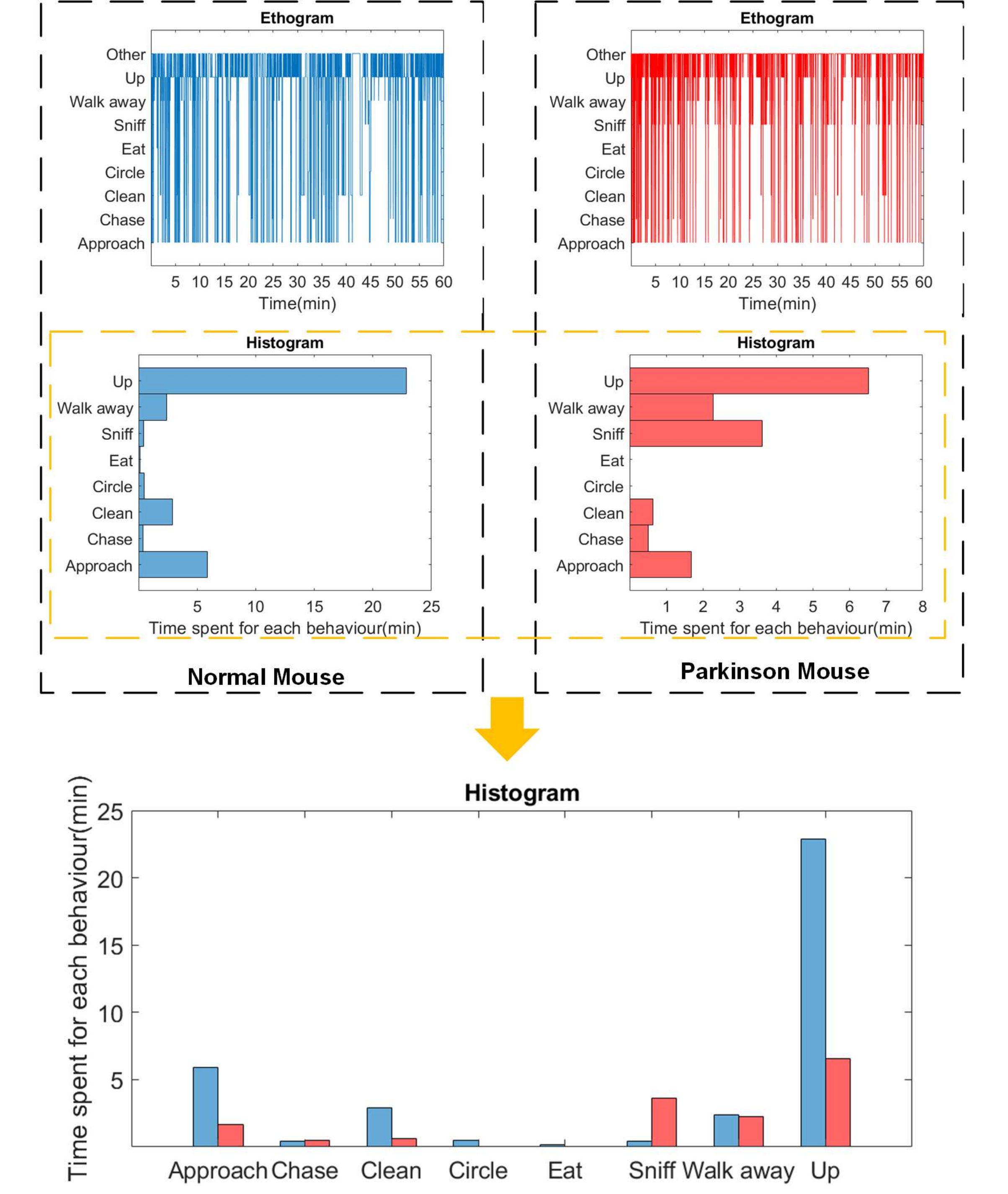}
\end{center}
\caption{Behaviour frequencies measured over the 60-min period for the mice with Parkinson's disease and their control strain. Blue - Control mice, and Red - Mice with Parkinson's disease. }
\label{fig:parkinson}
\end{figure}

To demonstrate the versatility of our proposed approach with different laboratory settings, as an example, we here use the proposed system to discriminate the behaviours of control mice and MPTP treated mice for Parkinson’s disease. Similar to CRIM13, the whole dataset is also evenly divided to training and testing datasets. Fig. S5 shows the agreement of the labeling results by our MV-LDCRF model and the expert annotators on the testing datasets. The agreement is satisfactory for most behaviours, whereas 18\% of the `approach' behaviour are incorrectly classified as ‘walk away’, 18\% of the `sniff' behaviour are incorrectly classified as ‘up’, and 16\% of the `up' behaviour are incorrectly classified as ‘walk away’. However, compared with the other methods, our approach still has the highest averaging accuracy 71.9\% and the best performance for 7 out of 8 behaviours, as shown in Table \ref{tab:our_dataset}. Experiments on both datasets have presented a high agreement rate by the proposed model. To demonstrate the applicability of the proposed system to behaviour phenotyping of the MPTP mouse model for Parkinson’s disease, we analyse the behaviour frequencies measured over the 60-min period for the MPTP treated mice and their control strains in Fig. \ref{fig:parkinson}. We observe that the MPTP treated mice, compared to the control group, have less exercises in ‘up’, ‘circle’, ‘clean’ and ‘approach’ and more exercises in ‘sniff’.

\begin{table*}
\begin{center}
\caption{Behaviour recognition results of various approaches for our own PDMB dataset.}
\label{tab:our_dataset}

\begin{tabular}{|p{1.4cm}|p{1.4cm}|p{1.4cm}|p{1.4cm}|p{1.4cm}|p{1.4cm}|p{1.4cm}|p{1.4cm}|p{1.4cm}|}
\hline
Behaviour & PBMV & KCCA (Gaussian) & DCCA & BILSTM & DCLSTM &LDCRF&MV-LDCRF& Ours (all)\\
\hline\hline
approach & 52.1\% & 55.6\% & 67.3\% & 67.6\% & 56.3\% &46.1\%&59.2\%& \textbf{68.0\%}\\
chase & 57.9\% & 73.4\% & 89.5\% & 67.9\% & 73.6\% &47.4\%&46.6\%& 78.9\%\\
circle & 54.2\% & 59.8\% & 75.6\% & 81.1\% & 73.8\% &73.1\%&76.3\%& \textbf{82.3\%}\\
eat & 42.3\% & 34.6\% & 49.8\% & 46.2\%& 50.3\% &51.7\%&63.1\%& \textbf{84.6\%}\\
clean & 25.0\% & 24.6\% & 25.4\% & 37.5\%& 25.4\% &49.3\%&56.3\%& \textbf{62.5\%}\\
sniff & 18.2\% & 31.8\% & 18.2\% & 4.6\% & 9.1\% &42.1\%&45.5\%& \textbf{63.6\%}\\
up & 21.1\% & 56.1\% & 15.5\% & 63.4\% & 64.2\% &64.0\%&59.4\%& \textbf{65.0\%}\\
walk away & 68.0\% & 70.7\% & 77.6\% & 82.1\% & 53.5\% &52.3\%&63.4\%&\textbf{86.0\%}\\
other & 80.5\% & 83.5\% & 89.1\% & 91.5\% & 89.2\% &75.4\%&77.8\%& 68.8\%\\
{\bf Average} & 46.6\% & 54.5\% & 56.4\% & 60.3\% & 57.3\% &55.7\%&60.8\%&\textbf{71.9\%}\\
\hline
\end{tabular}
\end{center}
\end{table*}

\section{Discussion and Conclusion}
Automated social behaviour recognition for mice is an important problem due to its clear benefits: repeatability, objectiveness, consistency, efficiency and cost-effectiveness. Traditional automated systems use sensors such as infrared sensors, radio-frequency identification (RFID) transponders and photobeams, and single 2D cameras. However, those sensor-based or single-view approaches restrict their abilities to recognise complex mouse behaviours. In contrast, multi-view behaviour recognition systems have demonstrated their potential to recognise mouse behaviours in occlusion.

In this paper, we have proposed a deep probabilistic model to perform multi-view social behaviour quantification in mice. Our approach jointly models the temporal relationship of frames in each view, the relationship between views and the correlation between labels in the neighbouring areas. Moreover, our system utilises both view-shared and view-specific features to accurately characterise mouse social behaviours in different scenarios.

We benchmarked every component of our approach separately. The performance of various feature extractors for mouse behaviour recognition was firstly evaluated on the CRIM13 dataset. The experimental results showed that the combination of TDD and VF\&CF had the highest AUC value and accuracy, outperforming the other combined features and the individual use of the features. This suggests that the multiplicity and complementarity of heterogeneous features provides very encouraging support in study of mouse behaviour. To verify the effectiveness of the view-shared substructure in our model, our system was tested independently and also compared to the other methods with view-shared feature representations. We showed that our approach achieved the best performance for 11 out of 12 behaviours. Thanks to variational inference, our model can effectively handle imbalanced data. Modelling label correlation has also been demonstrated to retain 6\% higher averaging accuracy than that without modelling label correlation. The statistical significance of our results was proved in our statistical analysis using two-sample t-test, paired t-test and Wilcoxon signed-rank tests. We also demonstrated that the performance of our approaches was not deteriorated over time. Compared to the other state-of-the-art methods that have the averaging accuracy of 62.6\%, our best model (with label correlation) achieved significantly better averaging accuracy of 71.7\%. A major advantage of our proposed method is that our model can automatically learn the contributions of each view-specific and view-shared feature, while the comparative approaches treat the features across views equally. On the other hand, we provided a new multi-view video dataset for motion monitoring of mice with Parkinson’s disease. We also validated our system on the PDMB dataset with two important aspects: the generalisation ability of the proposed deep graphical model on the new datasets and the applicability of the proposed system to behaviour phenotyping of the MPTP mouse model for Parkinson’s disease.

In addition, our experiments show that our spatio-temporal and trajectory-based motion features are still insufficient to distinguish between some similar behaviours such as `drink' and `eat', `approach' and `walk away'. We believe that it is possible to achieve better performance with (a) a better coordinated multi-camera system to share the visual information over views, and (b) development of characteristic features that capture mouse posture for mouse motion identification.

In summary, we describe the first deep graphical model, to our knowledge, of integrating features extracted from video recordings of multiple views, to perform automated quantification of social behaviours for freely interacting mice in a home-cage environment. The proposed approach has the potential to become a valuable tool for quantitative phenotyping of complex behaviours including those for the study of mice with neurodegenerative diseases. Furthermore, our approach can be potentially extended to other multi-view activity recognition, especially for the recognition of highly correlated behaviour in a long video recording over hours.

%\section*{Acknowledgement}
%This research was supported by UK EPSRC under Grant EP/N011074/1.

\ifCLASSOPTIONcaptionsoff
  \newpage
\fi

\bibliographystyle{IEEEtran}
\bibliography{mybibfile}
\begin{IEEEbiography}[{\includegraphics[width=1in,height=1.25in,clip,keepaspectratio]{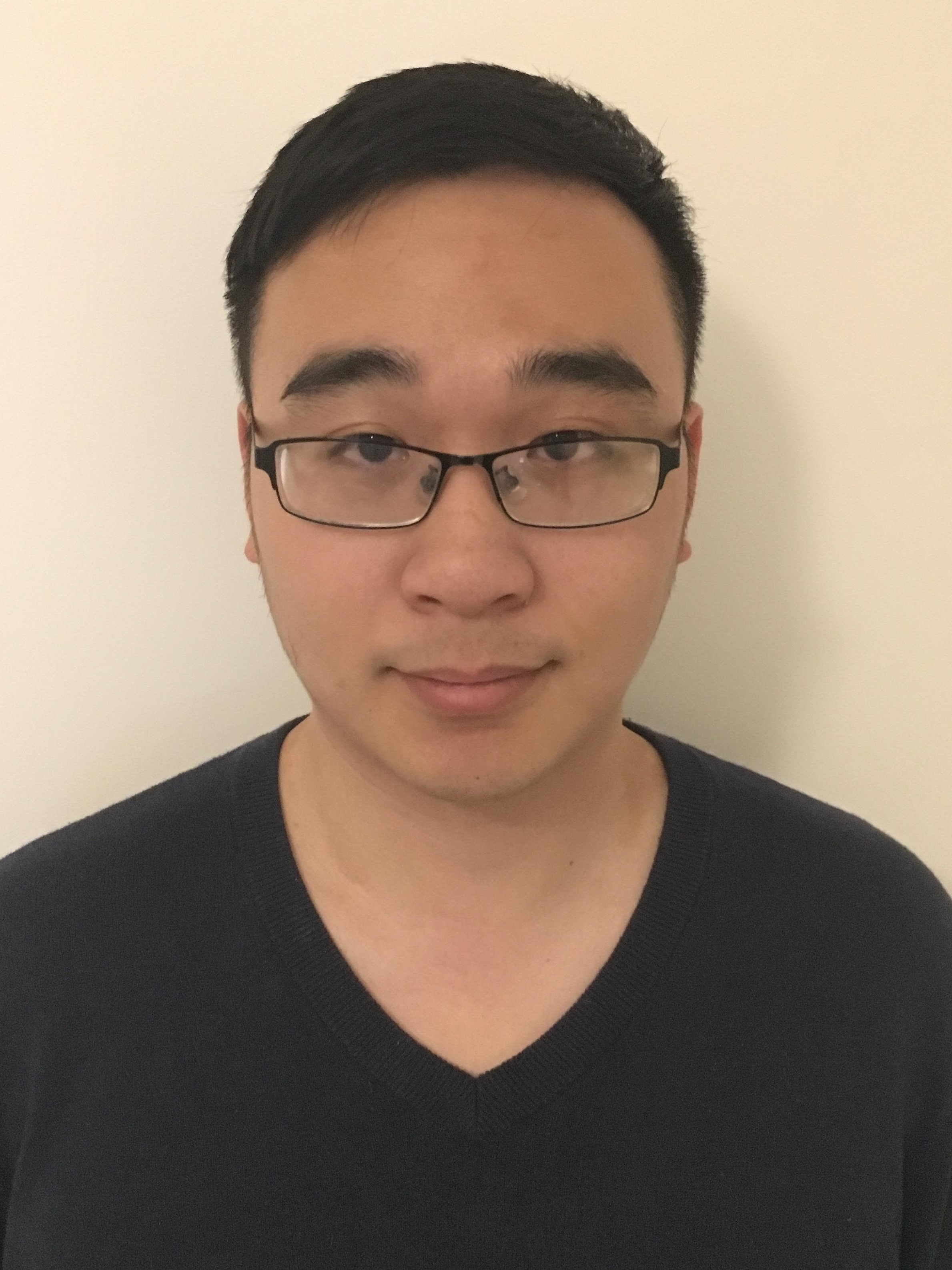}}]{Zheheng Jiang}
received the B.Sc. degree in Electrical Engineering and Automation (Grid Monitoring) from Nanjing Institute of Technology and the M.Sc. degree in Software Development from Queen’s University of Belfast, Belfast, U.K. He has been awarded his Ph.D. degree in Computer Science from University of Leicester, Leicester, U.K. He is currently the Senior Research Associate at the Computing and Communications, Lancaster University, Lancaster, U.K.

His current research interests include machine learning for vision, object detection and recognition, video analysis and event recognition.
\end{IEEEbiography}
\begin{IEEEbiography}[{\includegraphics[width=1in,height=1.25in,clip,keepaspectratio]{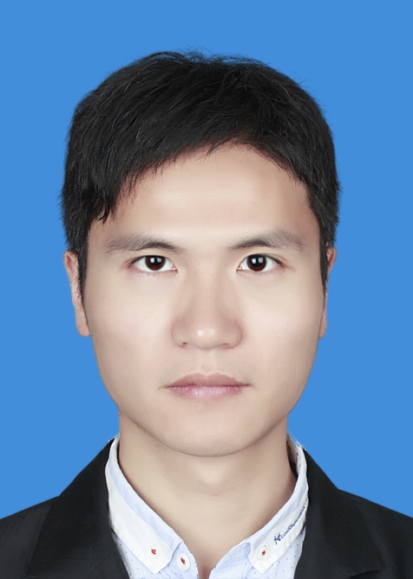}}]{Feixiang Zhou}received the B.S. degree in electronic science and technology from Changshu Institute of Technology, Suzhou, China, in 2016,  the M.S. degree in control theory and control engineering from Shanghai University, Shanghai, China, in 2019. He is currently pursuing the Ph.D. degree with the School of Informatics, University of Leicester, Leicester, U.K. 

His current research interests include Computer Vision, Machine Learning and their applications on video understanding.

\end{IEEEbiography}

\begin{IEEEbiography}[{\includegraphics[width=1in,height=1.25in,clip,keepaspectratio]{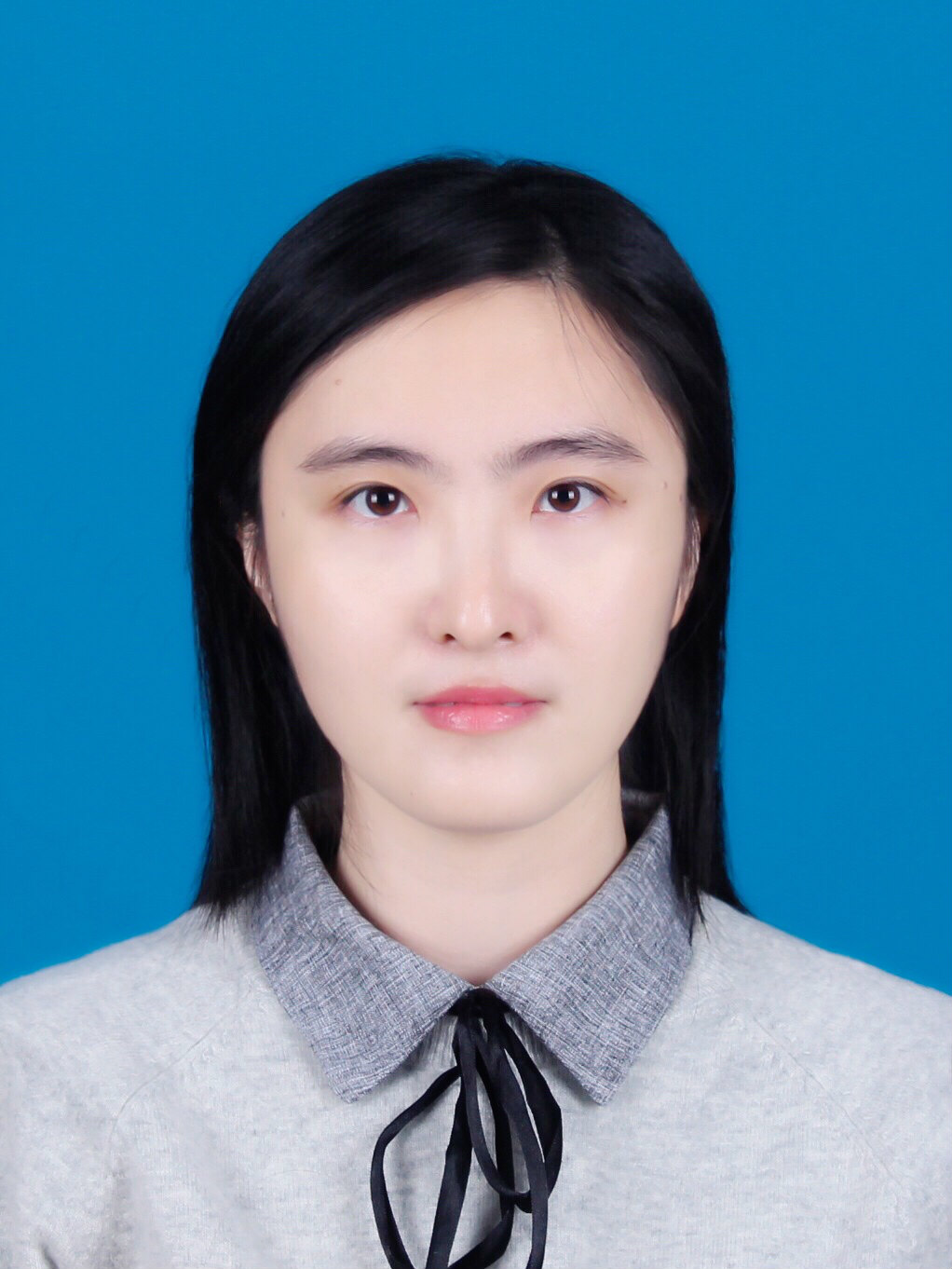}}]{Aite Zhao} received the Bachelor's degree in software engineering from Qingdao University of Technology in 2013, and received her Ph.D. degree in June 2020 in the College of Information Science and Engineering in Ocean University of China. She is a visiting Ph.D. researcher in the School of Informatics, University of Leicester, Leicester, U.K. She is currently a Lecturer of College of Computer Science and Technology in Qingdao University. 

Her research interests include computer vision, pattern recognition, machine learning, data analysis and robotics.

\end{IEEEbiography}

\begin{IEEEbiography}[{\includegraphics[width=1in,height=1.25in,clip,keepaspectratio]{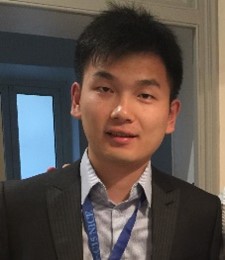}}]{Dr. Xin Li} obtained BEng in Electrical Information Engineering from the University of Science and Technology Beijing 2011 and MSc in Electrical Electronic Engineering from the University of Leeds in 2012. He has been awarded his PhD in Biomedical Engineering from the University of Leicester in 2016. He was appointed as Research Associate from 2016 and promoted to Lecturer in 2019 at Departments of Cardiovascular Sciences and Engineering, University of Leicester, UK.

His research focused on using advanced signal processing and mathematical intelligent algorithms for improving target identification for catheter ablation during human persistent atrial fibrillation and better risk assessment for sudden cardiac death.

\end{IEEEbiography}

\begin{IEEEbiography}[{\includegraphics[width=1in,height=1.25in,clip,keepaspectratio]{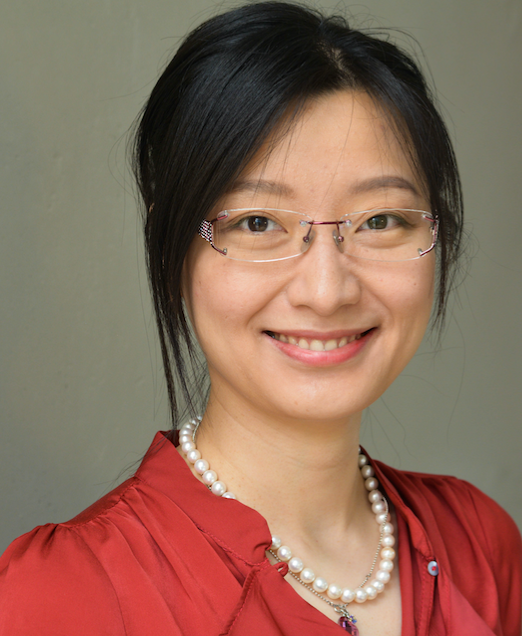}}]{Dr. Ling Li}
is the Director of Internationalisation at the School of Computing and also the founding coordinator of Laboratory of Brain $\mid$ Cognition $\mid$ Computing (BC2 Lab) of the school responsible for coordinating multidisciplinary research between Computing, Sports and local NHS hospitals. She had six-year research experience at Imperial College London with a focus to understand body sensor data (EEG, EMG, ECG, eAR-sensor, and etc.). She participated in large scale projects. She also involved in projects from government and industry (i.e. Samsung GRO award). She now serves at the editorial board of Brain Informatics and the secretary of IEEE Computing Society in UK and Ireland.
\end{IEEEbiography}

\begin{IEEEbiography}[{\includegraphics[width=1in,height=1.25in,clip,keepaspectratio]{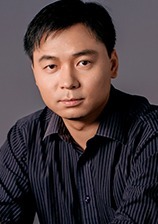}}]{Dacheng Tao (F'15)} ) is the Director of the JD Explore Academy and a Vice President of JD.com. He mainly applies statistics and mathematics to artificial intelligence and data science, and his research is detailed in one monograph and over 200 publications in prestigious journals and proceedings at leading conferences. He received the 2015 Australian Scopus-Eureka Prize, the 2018 IEEE ICDM Research Contributions Award, and the 2021 IEEE Computer Society McCluskey Technical Achievement Award. He is a fellow of the Australian Academy of Science, AAAS, ACM and IEEE.
\end{IEEEbiography}

\begin{IEEEbiographynophoto}{Xuelong Li(M'02-SM'07-F'12)}is a full professor with School of Artificial Intelligence, Optics and Electronics (iOPEN), Northwestern Polytechnical University, Xi'an 710072, P.R. China. 
 
\end{IEEEbiographynophoto}

\begin{IEEEbiography}[{\includegraphics[width=1in,height=1.25in,clip,keepaspectratio]{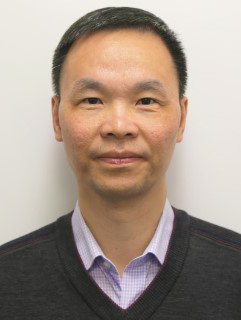}}]{Huiyu Zhou}
received a Bachelor of Engineering degree in Radio Technology from Huazhong University of Science and Technology of China, and a Master of Science degree in Biomedical Engineering from University of Dundee of United Kingdom, respectively. He was awarded a Doctor of Philosophy degree in Computer Vision from Heriot-Watt University, Edinburgh, United Kingdom. Dr. Zhou currently is a full Professor at School of Informatics, University of Leicester, United Kingdom. He has published over 350 peer-reviewed papers in the field.

His research work has been or is being supported by UK EPSRC, ESRC, AHRC, MRC, EU, Royal Society, Leverhulme Trust, Puffin Trust, Invest NI and industry.

\end{IEEEbiography}

\end{document}